# Model Change Active Learning in Graph-Based Semi-supervised Learning

**Kevin S. Miller**[1] · **Andrea L. Bertozzi**[1]

Received: 24 October 2022 / Revised: 18 September 2023 / Accepted: 21 September 2023 /
Published online: 17 February 2024

**Abstract**

Active learning in semi-supervised classification involves introducing additional labels for unlabelled data to improve the accuracy of the underlying classifier. A challenge is to identify which points to label to best improve performance while limiting the number of new labels. "Model Change" active learning quantifies the resulting change incurred in the classifier by introducing the additional label(s). We pair this idea with graph-based semi-supervised learning (SSL) methods, that use the spectrum of the graph Laplacian matrix, which can be truncated to avoid prohibitively large computational and storage costs. We consider a family of convex loss functions for which the acquisition function can be efficiently approximated using the Laplace approximation of the posterior distribution. We show a variety of multiclass examples that illustrate improved performance over prior state-of-art.

## 1 Introduction

Supervised machine learning algorithms rely on datasets that contain an abundance of labels (i.e., known classifications) for associated inputs. In many real-world applications, however, unlabeled data is ubiquitous, while obtaining labels for such training data is costly. Semi-supervised learning (SSL) methods leverage unlabeled data to achieve an accurate classification with significantly fewer training points. At the same time, the choice of training points often affects classifier performance, especially in the case of SSL due to the small training set size. Active learning seeks to choose a "useful" training set from

✉ Kevin S. Miller
millerk22@g.ucla.edu

Andrea L. Bertozzi
bertozzi@g.ucla.edu

[1] Department of Mathematics, University of California, Los Angeles, 520 Portola Plaza, Los Angeles, CA 90095, USA

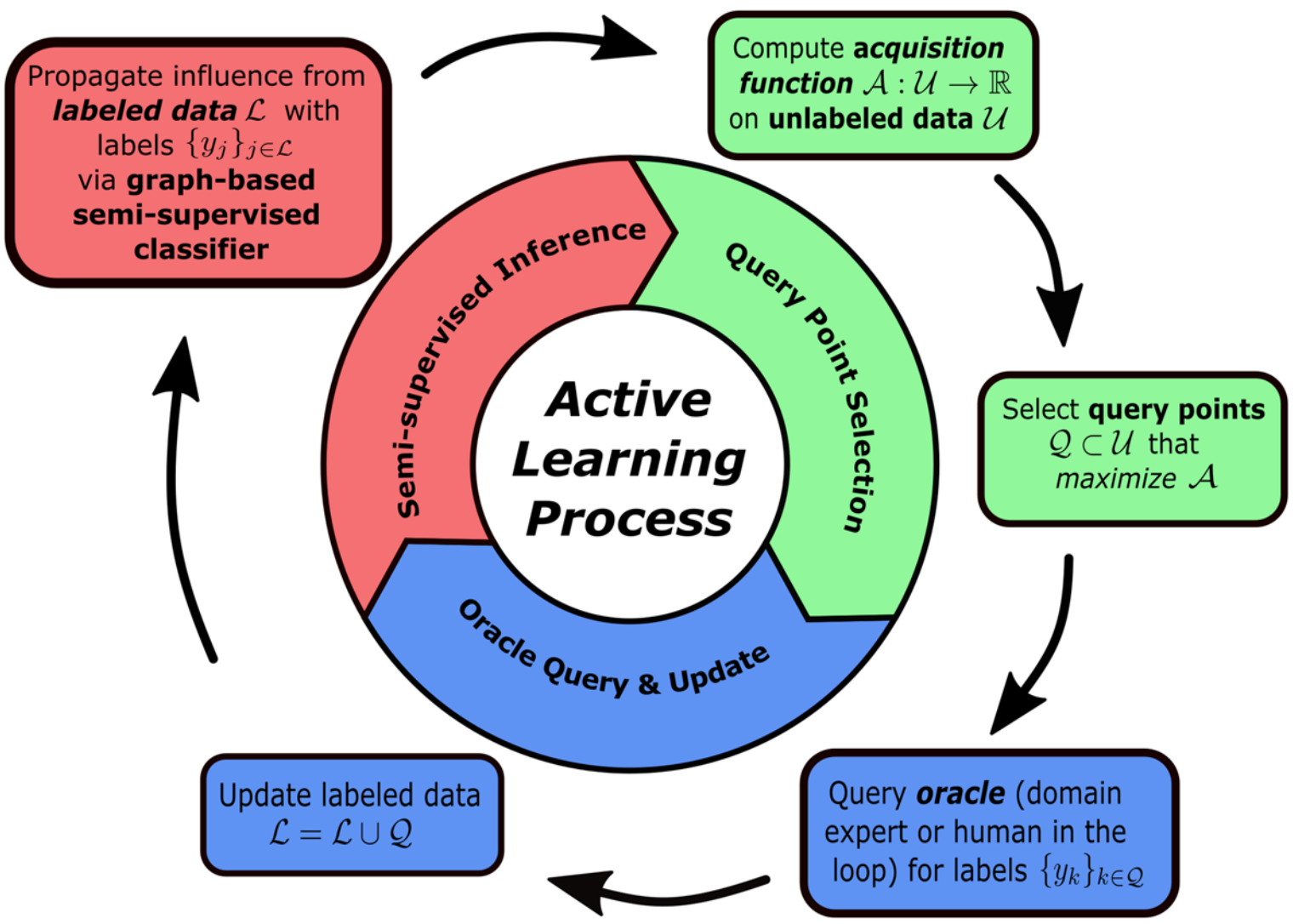

**Fig. 1** Active Learning Flowchart. The cycle begins by (1-red) training the underlying semi-supervised classifier with the current labeled set $\mathcal{L}$ with labels $\{y_j\}_{j\in\mathcal{L}}$. In this paper, we focus on graph-based SSL classifiers that propagate influence via a similarity graph (see Sect. 2.1). Query points $\mathcal{Q}\subset\mathcal{U}$ are then selected (2-green) from the unlabeled data ($\mathcal{U}$) as the maximizers of an *acquisition function* that quantifies the "informativeness" of each unlabeled point $k\in\mathcal{U}$ is evaluated. Finally, the query points are labeled (3-blue) according to an *oracle* (i.e., a domain expert or human in the loop labeler) and subsequently added to the labeled data $\mathcal{L}$. The process repeats with the updated labeled data, retraining the SSL classifier to prepare for another active learning query and update

which the underlying machine learning algorithm learns, and careful selection of such training data is motivated by the inherent cost to label data in practice. The main challenge in active learning is designing the criterion for selecting which unlabeled points are the most beneficial to label to significantly improve the underlying machine learning classifier's performance, more so than merely selecting random points to label. While there are a few different formulations of active learning, we focus on the *pool-based* active learning paradigm [35] as opposed to online or streaming-based active learning [48]. That is, the active learner has access to a fixed "pool" of unlabeled data points from which it can decide a subset to label.

In pool-based active learning, most methods alternate between: (i) training a model[1] given the current labeled data $\mathcal{L}$, $\{y_j\}_{j\in\mathcal{L}}$ and (ii) choosing a set of active learning query points $\mathcal{Q}$ in the unlabeled set $\mathcal{U}$ according to an *acquisition function* (also called an *active learning criterion*), see Fig. 1. An "oracle" that has access to the classification of all data then labels points $k\in\mathcal{Q}$; oftentimes in application, a domain expert plays the role of the oracle in identifying the true classification of points (i.e., "human-in-the-loop" schemes). We refer to this iterative procedure of alternating between model training and query set selection and labeling as the *active learning process*. The goal then is to design an acquisition function $\mathcal{A}\colon\mathcal{U}\to\mathbb{R}$ that quantifies how useful labeling a point $k\in\mathcal{U}$ would be in the active learning process. With a specified acquisition function $\mathcal{A}$, we select the query set $\mathcal{Q} \subset \mathcal{U}$ in the active learning process either *sequentially* (i.e., one at a time, $|\mathcal{Q}| = 1$) or in a *batch* (i.e., $|\mathcal{Q}| = B \in \mathbb{N}_+$).

[1] While "model training" in deep learning models usually requires non-convex optimization of model parameters, the graph-based models we consider in this work are "trained" by simply solving a convex optimization problem that can also be efficiently approximated.

Most active learning acquisition functions belong to one of a few categories: uncertainty [19, 25, 48], margin [2, 28, 50], clustering [16, 39], and look-ahead [10, 56]. *Uncertainty*-based acquisition functions favor unlabeled points whose classification is "uncertain". Criterion such as entropy [48], least-confident [48], Query by Disagreement (QBD) [2, 14], and Bayesian Active Learning by Disagreement (BALD) [25] fall into this category. Closely related to uncertainty-based acquisition functions are *margin*-based acquisition functions that favor unlabeled points near the decision boundary of the current SSL classifier [24, 28, 50]. Support Vector Machines (SVM) [51] are amenable to this type of criterion, as the concept of a margin and decision boundary are inherent in the model. *Clustering*-based methods rely on the geometric clustering structure of the input data to guide the active learning query choices. The works of Dasgupta and Hsu [16], Dasarathy et al. (S2) [15], and Maggioni and Murphy [39] are examples of acquisition functions that specifically exploit the clustering structure of the input data, usually based on a graph topology that captures these geometric relationships.

*Look-ahead* acquisition functions are a final category that we mention, and are the motivation for the present work. Look-ahead methods leverage the current SSL classifier's state to "look ahead" at what changes would occur in the SSL classifier as a result of labeling an unlabeled point, such as the seminal work of Zhu et al. [56]. Our proposed "Model Change" acquisition function as well as the EMCM [10] and Maxi-Min "data-based norm" [30] methods is in this category. More specifically, the proposed Model Change acquisition function $\mathcal{A}(k)$ of this paper approximates the difference between the *current* classifier $\mathbf{u}$ and the *look-ahead* classifier $\mathbf{u}^{+k,\hat{y}_k}$ resulting from adding $k$ to the labeled set $\mathcal{L}$ with a hypothetical label, $\hat{y}_k$.

While many methods of late focus on applying active learning in deep learning architectures [1, 19, 33, 47, 49], we focus on graph-based SSL models in this paper. Graph-based SSL models leverage the geometric information from a similarity graph imposed on the set of feature vectors in conjunction with the previously observed labeling information contained in the labeled set, $\mathcal{L} \subset Z$. These models allow for straightforward Bayesian probabilistic interpretations that are less clear in the majority of deep learning architectures [19].

The contributions of this work are to

- provide a unifying framework for active learning in graph-based SSL models with convex loss functions more appropriate for classification tasks;
- present "Model Change" active learning acquisition function built around fast look-ahead approximations previously only performed on interpolation RKHS models [30];
- apply a spectral truncation to the graph-based SSL models which allows for efficient storage and Model Change calculations;
- demonstrate the speed and efficacy of the approach on both synthetic and real-world datasets, including an application to hyperspectral imagery (HSI).

## 2 Background

We first introduce the family of graph-based SSL models that use the Model Change acquisition function. Then we discuss the probabilistic counterpart to these graph-based SSL models and how these Bayesian perspectives arise in previous work in active learning.

### 2.1 Graph-Based SSL Models

Consider the input data $X = \{\mathbf{x}_1, \mathbf{x}_2, \cdots, \mathbf{x}_N\} \subset \mathbb{R}^d$ of $N$ feature vectors with a corresponding index set $\mathcal{Z} = \{1, 2, \cdots, N\}$. Assume there exists a "ground-truth" classification (labeling) function $y^\dagger : \mathcal{Z} \to \{1, 2, \cdots, n_c\}$ that maps each point $\mathbf{x}_i$ to exactly one class, identified by the label $y_i^\dagger \in \{1, 2, \cdots, n_c\}$[2]. As we are assuming the semi-supervised case of learning, we have *observations* $y_j$ of these ground-truth labelings for only a subset of indices $j \in \mathcal{L} \subset Z$, i.e., the labeled set. The goal of SSL is to infer the ground truth classifications $y_k^\dagger$ for $k \in \mathcal{U} = \mathcal{Z} - \mathcal{L}$, i.e., the unlabeled set. In the binary case, we denote the concatenation of the observed labelings $\{y_j\}_{j\in\mathcal{L}}$ as a vector $\mathbf{y} \in \mathbb{R}^{|\mathcal{L}|}$, whereas in the multiclass case we denote the concatenations of the observed one-hot vectors $\{\mathbf{y}_j\}_{j\in\mathcal{L}}$ as a matrix $\mathbf{Y} \in \mathbb{R}^{|\mathcal{L}|\times n_c}$. For practicality, assume that the observed labels $\{y_j\}_{j\in\mathcal{L}}$ have *at least one* label per class; that is, $\cup_{j\in\mathcal{L}} y_j = \{1, 2, \cdots, n_c\}$ in the multiclass case (or $\{\pm 1\}$ in the binary case)[3].

We construct a similarity graph $G(\mathcal{Z}, \mathbf{W})$ with edge weights $\mathbf{W}_{ij} = \kappa(\mathbf{x}_i, \mathbf{x}_j) \geqslant 0$ calculated by a similarity kernel $\kappa$. For example, the Gaussian kernel, $\kappa(\mathbf{x}_i, \mathbf{x}_j) = \exp(-\|\mathbf{x}_i - \mathbf{x}_j\|_2^2/\sigma^2)$ with kernel width parameter $\sigma > 0$, is a popular choice. Important geometric information about the data manifold is encoded in graph Laplacian matrices [3, 52]. Two such matrices are the *unnormalized graph Laplacian matrix* $\mathbf{L}_u = \mathbf{D} - \mathbf{W}$ and the *normalized graph Laplacian matrix* $\mathbf{L} = \mathbf{D}^{-1/2}(\mathbf{D} - \mathbf{W})\mathbf{D}^{-1/2}$, where $\mathbf{D} = \mathrm{diag}(d_1, d_2, \cdots, d_N), d_i = \sum_{j\neq i} \mathbf{W}_{ij}$ is the diagonal *degree matrix*. There are other possible graph Laplacian matrices one could define, but we only consider the normalized graph Laplacian, $\mathbf{L}$. We enforce symmetry in $\mathbf{L}$ to ensure that the eigenvalues and eigenvectors of this matrix are real-valued. Graph Laplacian matrices are positive semi-definite and so to ensure the invertibility of this matrix we consider the perturbation $\mathbf{L}_\tau = \mathbf{L} + \tau^2\mathbf{I}$ with parameter $\tau > 0$. The matrix $\mathbf{L}_\tau$ is now positive definite and invertible, which also ensures a well-defined Bayesian prior distribution that we present in Sect. 2.2.

**Remark 1** The optimal selection of $\tau$ is subtle and not rigorously well-understood. In the context of posterior consistency of a Bayesian probabilistic of graph-based semi-supervised regression, the authors in [6] showed that $\tau$ should be selected with respect to the "well-clusteredness" of the dataset. This is not an explicitly known quantity a priori, though it is reasonable to choose $\tau$ in relation to the *spectral gap* of the graph Laplacian. While this is an interesting line of inquiry, it is out of the scope of the present work and so we leave this for future work. In the numerical results of Sect. 4, we set a fixed choice of $\tau = 0.001, 0.005$ to allow for consistency across experiments, similar to that done in other works with the Harmonic Functions model [29, 56].

Define a continuous-valued function on the nodes of the graph $u\colon \mathcal{Z} \to \mathbb{R}^{n_c}$. This function can be identified with a matrix

[2] In the binary case we take $y_i^\dagger \in \{\pm 1\}$.

[3] It is an interesting question to consider how well active learning methods can explore "undiscovered" classes (i.e., $\cup_{j\in\mathcal{L}} y_j \neq \{1, 2, \cdots, n_c\}$), but we leave this for future work.

$$\mathbf{U} = \begin{pmatrix} \mathbf{u}_1 \\ \mathbf{u}_2 \\ \vdots \\ \mathbf{u}_N \end{pmatrix} \in \mathbb{R}^{N \times n_c},$$

where the $i^{th}$ row of $\mathbf{U}$, $\mathbf{u}_i$, corresponds to the inferred classification of node $i$ via a mapping $S\colon \mathbb{R}^{n_c} \to \{1, 2, \cdots, n_c\}$ that maps the vectors $\{\mathbf{u}_i\}_{i \in Z}$ to the space of possible classes. For example, the mapping $S(\mathbf{u}_i) = \underset{c=1,2,\cdots,n_c}{\arg\max}\ [\mathbf{u}_i]_c$ infers the classification of node $i$ as the index of the maximal element of the vector. Other thresholding functions have been explored for providing consistent classifiers [11]. In the special case of binary classification, the matrix $\mathbf{U}$ actually can be represented by a vector $\mathbf{u} \in \mathbb{R}^N$, with mapping $S(u_i) = \mathrm{sgn}(u_i) = \delta_{\{u_i \geqslant 0\}}$.

Graph-based SSL then leverages the graph Laplacian matrix via a quadratic form to regularize the SSL problem in order to encourage similar labels for inputs that lie close to each other on the underlying data manifold [3]. In the binary case, we solve the following optimization problem:

$$\hat{\mathbf{u}} = \underset{\mathbf{u} \in \mathbb{R}^N}{\arg\min}\ \frac{1}{2}\langle \mathbf{u}, \mathbf{L}_\tau \mathbf{u} \rangle + \sum_{j \in \mathcal{L}} \ell(u_j, y_j) =: \underset{\mathbf{u} \in \mathbb{R}^N}{\arg\min}\ J_\ell(\mathbf{u}; \mathbf{y}), \tag{1}$$

where $\ell\colon \mathbb{R} \times \mathbb{R} \to [0, \infty)$ is a chosen loss function and $y_j \in \{\pm 1\}$. In the multiclass case, the objective function and corresponding inference $\hat{\mathbf{U}}$ are

$$\hat{\mathbf{U}} = \underset{\mathbf{U} \in \mathbb{R}^{N \times n_c}}{\arg\min}\ \frac{1}{2}\langle \mathbf{U}, \mathbf{L}_\tau \mathbf{U} \rangle_{\mathrm{F}} + \sum_{j \in \mathcal{L}} \ell(\mathbf{u}_j, \mathbf{y}_j) =: \underset{\mathbf{U} \in \mathbb{R}^{N \times n_c}}{\arg\min}\ \mathcal{J}_\ell(\mathbf{U}; \mathbf{Y}), \tag{2}$$

where $\ell\colon \mathbb{R}^{n_c} \times \mathbb{R}^{n_c} \to [0, \infty)$ is a chosen loss function and $\langle \cdot, \cdot \rangle_{\mathrm{F}}$ is the Frobenius inner product.

Table 1 lists a family of graph-based SSL models defined by the objective functions of (1) and (2); the Model Change acquisition function of this present work applies to each. Graph-based SSL models historically have been motivated by Gaussian Random Fields (GRF) [55], and the *Harmonic Functions* (HF) model [56] has been influential in active learning in graph-based SSL. This model can be thought of using a "hard-constraint" loss function, where $\ell(x, y) = 0$ if $x = y$ and $+\infty$ otherwise. Acquisition functions minimizing look-ahead expected risk (MBR [56], TSA [29]), posterior variance (V-Opt [26], Σ-Opt [37]), and other measures of uncertainty [36] come from this model. The authors in [33] even used the HF model to enhance active learning performance within deep learning.

### 2.1.1 Cross-Entropy (CE) Model

As an alternative to regression for the multiclass setting, we show how to incorporate the cross-entropy (CE) loss function $\ell(\mathbf{x}, \mathbf{y}) = -\sum_{c=1}^{n_c} x_c \ln(y_c)$ into the multiclass graph-based SSL framework (2). The CE loss requires that both inputs are probability distribution vectors on the set of possible classes, $\{1, 2, \cdots, n_c\}$. While the observations $\mathbf{y}_j$ satisfy this property because of their one-hot form, the rows $\mathbf{u}_i$ of arbitrary $\mathbf{U} \in \mathbb{R}^{N \times n_c}$ do not necessarily satisfy this same condition. The entries of $\mathbf{U}$ are not even constrained to be non-negative. As such, we apply the"softmax" function on the rows of $\mathbf{U}$ to enforce this probability distribution property. Denoting the $(j, c)^{th}$ entry of $\mathbf{U}$ by $\mathbf{U}_{j,c}$ and the $c^{th}$ entry of $\mathbf{y}_j$ by $[\mathbf{y}_j]_c$, we have

$$\mathcal{S}(\mathbf{u}_j; \gamma) := \frac{1}{\sum_{h=1}^{n_c} \mathrm{e}^{\mathbf{U}_{j,h}/\gamma}} (\mathrm{e}^{\mathbf{U}_{j,1}/\gamma}, \mathrm{e}^{\mathbf{U}_{j,2}/\gamma}, \cdots, \mathrm{e}^{\mathbf{U}_{j,n_c}/\gamma})^{\mathrm{T}},$$

so that the loss function can be written as

$$\begin{aligned}\sum_{j\in\mathcal{L}} \ell(\mathbf{y}_j, \mathbf{u}_j) &= -\sum_{j\in\mathcal{L}} \sum_{c=1}^{n_c} [\mathbf{y}_j]_c \ln\left([\mathcal{S}(\mathbf{u}_j; \gamma)]_c\right) \\ &= \sum_{j\in\mathcal{L}} \left( -\frac{1}{\gamma}\langle \mathbf{y}_j, \mathbf{u}_j\rangle + \ln\left(\sum_{h=1}^{n_c} \mathrm{e}^{\mathbf{U}_{j,h}/\gamma}\right)\right),\end{aligned}$$

recalling the one-hot vector $\mathbf{y}_j = \mathbf{e}_{y_j} \in \mathbb{R}^{n_c}$. We write the CE model as

$$\mathcal{J}_{\mathrm{CE}}(\mathbf{U}; \mathbf{Y}) = \frac{1}{2}\langle \mathbf{U}, \mathbf{L}_\tau \mathbf{U}\rangle_{\mathrm{F}} + \sum_{j\in\mathcal{L}} \left( -\frac{1}{\gamma}\langle \mathbf{y}_j, \mathbf{u}_j\rangle + \ln\left(\sum_{h=1}^{n_c} \mathrm{e}^{\mathbf{U}_{j,h}/\gamma}\right)\right). \tag{3}$$

**Table 1** Family of graph-based SSL models. We extend the results of MC [41] to be more computationally efficient and allow for multiclass classification

| GBSSL model name | $\ell(x, y)$ | Prev. Acq. F'ns |
|---|---|---|
| Harmonic functions (HF) [55] | Hard-constraint | MBR [56], V-Opt [26], Σ-Opt [37], TSA [29], Entropy [36] |
| Gaussian regression (GR) [38] | $\frac{1}{2\gamma^2}(x-y)^2$ | Σ-Opt [38], MC [41], this work |
| Logistic (LOG) [23] | $\ln\left(1+\mathrm{e}^{-xy/\gamma}\right)$ | MC [41], this work |
| Probit (P) [23, 46] | $-\ln \Psi_\gamma(xy)$[a] | UQ-Sampling [7], MC [41], this work |
| Multiclass Gaussian Regression (MGR) [6] | $\frac{1}{2\gamma^2}\Vert\mathbf{x}-\mathbf{y}\Vert_2^2$ | This work |
| Cross-Entropy (CE) [27, 31] | $-\sum_{c=1}^{n_c} x_c \ln(y_c)$[b] | This work |

[a] $\Psi_\gamma(t) = \int_{-\infty}^{t} \psi_\gamma(s)\mathrm{d}s$ is the cumulative distribution function (CDF) of a log-concave probability density function (PDF) $\psi_\gamma(s)$ with parameter $\gamma > 0$

[b] This loss requires that $\mathbf{x}, \mathbf{y}$ be discrete probability distributions, which is adapted in Sect. 2.1.1

## 2.2 Bayesian Interpretation of SSL Problems

These graph-based SSL objective functions lend themselves to a Bayesian probabilistic interpretation, as discussed in prior literature [7, 23, 26, 38, 41, 45, 56]. In the binary case, (1) is equivalent to finding the *maximum a posteriori* (MAP) estimate of a posterior probability distribution whose density $\mathbb{P}(\mathbf{u}|\mathbf{y})$ relates to the objective function via

$$\mathbb{P}(\mathbf{u}|\mathbf{y}) \propto \exp\left(-J_\ell(\mathbf{u}; \mathbf{y})\right) = \mathrm{e}^{-\frac{1}{2}\langle \mathbf{u}, \mathbf{L}_\tau \mathbf{u}\rangle} \mathrm{e}^{-\sum_{j\in\mathcal{L}} \ell(u_j, y_j)} \propto \mu(\mathbf{u}) \mathrm{e}^{-\Phi_\ell(\mathbf{u}; \mathbf{y})}, \tag{4}$$

where the *prior* $\mu(\mathbf{u})$ follows a Gaussian prior, $\mathcal{N}(\mathbf{0}, \mathbf{L}_\tau^{-1})$, and the *likelihood*, $q(\mathbf{y}|\mathbf{u}) \propto \exp(-\Phi_\ell(\mathbf{u}; \mathbf{y}))$, is defined by the likelihood potential $\Phi_\ell(\mathbf{u}; \mathbf{y}) := \sum_{j\in\mathcal{L}} \ell(u_j, y_j)$. This prior relates to other graph-based SSL priors proposed in previous literature [23]. The Gaussian prior represents a prior belief over the distribution of functions $\mathbf{u}$ on the nodes of the graph, per the geometry of the data captured in the graph Laplacian matrix. The likelihood represents assumptions about the generative model that created the observed labelings $y_j$ from the ground-truth classifications $y_j^\dagger$. Each loss function $\ell$ for (1) defines a different likelihood and consequently a different modeling assumption about the observed labels $\mathbf{y}$ (or $\mathbf{Y}$). We note that although the prior $\mu(\mathbf{u})$ is Gaussian, the posterior distribution of (4) for general loss functions $\ell$ is not necessarily Gaussian.

A key idea of the present work is to approximate the non-Gaussian posterior distributions via suitable Gaussian distributions to exploit the resulting convenient form of what we term the "look-ahead" posterior mean and covariance. This formulation allows us to use loss functions that are arguably more natural for classification than merely the squared-error loss.

## 2.3 Look-Ahead Model

Recalling the binary graph-based SSL model objective

$$J_\ell(\mathbf{u};\mathbf{y}) = \frac{1}{2}\langle \mathbf{u}, \mathbf{L}_\tau \mathbf{u}\rangle + \sum_{j\in\mathcal{L}} \ell(u_j, y_j),$$

then let the "look-ahead" model objective be

$$J_\ell^{k,\hat{y}_k}(\mathbf{u};\mathbf{y},\hat{y}_k) := \frac{1}{2}\langle \mathbf{u}, \mathbf{L}_\tau \mathbf{u}\rangle + \sum_{j\in\mathcal{L}} \ell(u_j, y_j) + \ell(u_k,\hat{y}_k) = J_\ell(\mathbf{u};\mathbf{y}) + \ell(u_k,\hat{y}_k),$$

where we add the unlabeled index $k \in \mathcal{U}$ with pseudo-label $\hat{y}_k := \mathrm{sgn}(\hat{u}_k)$ to the labeled data. Look-ahead models have previously been introduced for designing various acquisition functions (e.g., [9, 10, 30, 56]). We emphasize here that in application, since $k \in \mathcal{U}$, we do not have access to the *true* labeling $y_k^\dagger$, and so this look-ahead model is a *hypothetical model*. For a given $k \in \mathcal{U}$ and one-hot encoding $\hat{\mathbf{y}}_k \in \mathbb{R}^{n_c}$ of the pseudo-label $\hat{y}_k = S(\hat{\mathbf{u}}^k) \in \{1, 2, \cdots, n_c\}$, the multiclass look-ahead model becomes

$$\begin{aligned}\mathcal{J}_\ell^{k,\hat{\mathbf{y}}_k}(\mathbf{U};\mathbf{Y},\hat{\mathbf{y}}_k) &:= \frac{1}{2}\langle \mathbf{U}, \mathbf{L}_\tau \mathbf{U}\rangle_{\mathrm{F}} + \sum_{j\in\mathcal{L}} \ell(\mathbf{u}_j, \mathbf{y}_j) + \ell(\mathbf{u}_k,\hat{\mathbf{y}}_k)\\ &= \mathcal{J}_\ell(\mathbf{U};\mathbf{Y}) + \ell(\mathbf{u}_k,\hat{\mathbf{y}}_k).\end{aligned}$$

The present work exploits a key property of the HF, GR, and MGR models that the *look-ahead* model's posterior mean (i.e., the graph-based SSL model's classifier) and covariance matrix are easily calculated as rank-one updates of the *current* model's posterior mean and covariance matrix. We discuss this more in Sects. 3.2, 3.3, and 3.4.

## 2.4 Laplace Approximation

*Laplace approximation* is a popular technique for approximating non-Gaussian distributions with a Gaussian distribution [46]. A given probability distribution, identified by its probability density function (PDF) $\mathbb{P}(\mathbf{x})$, can be approximated via the Gaussian distribution

$$\mathbf{x} \sim \mathcal{N}(\hat{\mathbf{x}}, \hat{\mathbf{C}}), \quad \hat{\mathbf{x}} = \underset{\mathbf{x}\in\mathbb{R}^N}{\arg\max}\ \mathbb{P}(\mathbf{x}), \quad \hat{\mathbf{C}} = \left(-\nabla^2 \ln(\mathbb{P}(\mathbf{x}))|_{\mathbf{x}=\hat{\mathbf{x}}}\right)^{-1},$$

where $\hat{\mathbf{x}}$ is the MAP estimator of $\mathbb{P}$ and $\hat{C}$ is the Hessian matrix of the negative-log density of the distribution evaluated at the MAP estimator, $\hat{\mathbf{x}}$. By applying the Laplace approximation to the non-Gaussian posterior distributions of Sect. 2.2, one can approximate look-ahead updates for calculating the Model Change acquisition function.

### 2.5 Active Learning Query Set Selection

With acquisition function $\mathcal{A}(k)$ in hand, one must select the query set $\mathcal{Q} \subset \mathcal{U}$ from these acquisition values. *Sequential* active learning selects the maximizer $k^* = \underset{k\in\mathcal{U}}{\arg\max}\ \mathcal{A}(k)$ of the acquisition function (i.e., $|\mathcal{Q}| = 1$). In *batch* active learning ($|\mathcal{Q}| = B > 1$), there is an added difficulty of how to best choose this subset. Choosing the top $B$ maximizers of the *current* values of $\{\mathcal{A}(k)\}_{k\in\mathcal{U}}$ can yield sub-optimal results, as these maximizers often are close in the ambient feature space—in a sense "wasting" precious query budget on redundant information. Some batch active learning methods select query points via a greedy sequential process [22, 26, 32, 37, 42], wherein the method selects the maximizer $k_1^* = \underset{k\in\mathcal{U}}{\arg\max}\,\mathcal{A}(k)$ first, and then updates the acquisition function values per the new, hypothetical SSL model with added index $k_1^*$ and associated pseudolabel $\hat{y}_{k_1^*}$. They next select the maximizer of the updated acquisition values $k_2^* = \underset{k\in\mathcal{U}-\{k_1^*\}}{\arg\max}\,\mathcal{A}^{k_1^*,\hat{y}_{k_1^*}}(k)$, and continue likewise to select $k_1^*, k_2^*, \cdots, k_B^*$. These methods frame batch query selection in terms of *submodular set functions*, wherein the subset chosen from the greedy sequential selection process is guaranteed to be near optimal [43]. Our presented method does not satisfy the properties of submodular set functions and so we turn to an alternative method for applying batch active learning.

Other batch active learning methods restrict the size of the set of indices on which $\mathcal{A}$ is evaluated to a smaller *candidate set* $\mathcal{S} \subset \mathcal{U}$. These methods select the batch $\mathcal{Q} \subset \mathcal{S}$ to be the top maximizers of the acquisition function [19, 33], where $\mathcal{S}$ is chosen uniformly at random form $\mathcal{U}$. This has essentially two important and positive consequences. First, evaluating $\mathcal{A}$ only over $\mathcal{S}$ is obviously computationally faster since $|\mathcal{S}| \ll |\mathcal{U}|$. Second, by selecting $\mathcal{S} \subset \mathcal{U}$ at random, we partially alleviate the problem of "redundant" calculations since the maximizers of $\mathcal{A}$ over $\mathcal{S}$ likely do not lie all close together. We apply this query set selection method to our batch active learning experiments; i.e., select $\mathcal{S} \subset \mathcal{U}$ uniformly at random and then select the top $B$ maximizers of the acquisition function.

## 3 Model Change Acquisition Derivation

We derive the Model Change acquisition function for a modified objective function that utilizes only a subset of the eigenvalues and eigenvectors of the graph Laplacian matrix. Section 3.1 defines spectral truncation modification of the family of graph-based SSL objectives. We derive in Sect. 3.2 the Model Change acquisition function first for the binary classification models by utilizing the Laplace approximation and a corresponding simple approximate update via Newton's method. We expand this idea to multiclass models in Sects. 3.3 and 3.4.

### 3.1 Spectral Truncation

Bayesian-inspired graph-based acquisition functions often require storing a prohibitively large and dense covariance matrix $\hat{\mathbf{C}} \in \mathbb{R}^{N\times N}$ in memory, where $N$ is the size of the dataset. We introduce a modification to the graph-based models to alleviate this burden by using only a subset of the eigenvalues and eigenvectors of the corresponding graph Laplacian matrix $\mathbf{L}$; we refer to this as "spectral truncation". We note that this form of spectral truncation is all but necessary for applying similarity graph-based methods that utilize covariance matrices related to the graph Laplacian matrix $\mathbf{L}$ to large-scale datasets due to the large memory footprint. Previous works [40, 44] have utilized techniques such as Nyström method [18, 53] for computing low-rank approximations of the (pseudo)inverse of graph Laplacians. In addition to the reduced memory footprint, spectral truncation also allows for time-efficient computations of the Model Change acquisition function.

The eigenvalues of the graph Laplacian matrix $\mathbf{L}$ can be ordered $0 = \lambda_1 \leqslant \lambda_2 \leqslant \cdots \leqslant \lambda_N$, with the first eigenvalue guaranteed to be 0 by properties of $\mathbf{L}$ [52]. The low-lying eigenvalues (i.e., closer to 0) and their corresponding eigenvectors of $\mathbf{L}$ contain important geometric information pertaining to the data. For example, spectral clustering [52] utilizes the eigenvectors corresponding to the first $K$ eigenvalues of $\mathbf{L}$ to embed the data into $\mathbb{R}^K$ and thereafter perform clustering, often via the K-Means algorithm. It was shown in [21] that for well-clustered data containing $K$ clusters, this corresponding "graph Laplacian embedding" of nodes for spectral clustering forms an orthogonal cone structure that highlights the primary clustering structure of the dataset. While spectral clustering is an unsupervised learning method, the same motivation of capturing the geometric clustering structure via graph Laplacian matrices is common to other graph-based SSL methods that use a spectral truncation [5, 8, 40].

Motivated by the previous discussion, the present work uses the $M$ eigenvalues closest to 0 and their corresponding eigenvectors to obtain approximations of $\hat{\mathbf{C}}$ and $\hat{\mathbf{u}}$ ($\hat{\mathcal{C}}$ and $\hat{\mathbf{U}}$). Recalling the perturbed graph Laplacian, $\mathbf{L}_\tau = \mathbf{L} + \tau^2\mathbf{I}$, then by considering only the first $M < N$ eigenvalues of $\mathbf{L}$, define the matrices

$$\mathbf{\Lambda}_\tau = \mathrm{diag}\big(\lambda_1 + \tau^2, \lambda_2 + \tau^2, \cdots, \lambda_M + \tau^2\big), \quad V = [\mathbf{v}^1, \mathbf{v}^2, \cdots, \mathbf{v}^M] \in \mathbb{R}^{N\times M},$$

where $\mathbf{v}^i$ is the eigenvector corresponding to the $i^{th}$ eigenvalue $\lambda_i$. Replace the graph-based regularization terms of (1) and (2) to obtain the following:

$$\begin{aligned} J_\ell(\mathbf{u};\mathbf{y}) &= \frac{1}{2}\langle \mathbf{u}, \mathbf{L}_\tau \mathbf{u}\rangle + \sum_{j\in\mathcal{L}} \ell(u_j, y_j) \\ &\to \frac{1}{2}\langle \boldsymbol{\alpha}, \mathbf{\Lambda}_\tau \boldsymbol{\alpha}\rangle + \sum_{j\in\mathcal{L}} \ell(\mathbf{e}_j^{\mathrm{T}} V\boldsymbol{\alpha}, y_j) =: \tilde{J}_\ell(\boldsymbol{\alpha};\mathbf{y}), \end{aligned} \tag{5}$$

$$\begin{aligned} \mathcal{J}_\ell(\mathbf{U};\mathbf{Y}) &= \frac{1}{2}\langle \mathbf{U}, \mathbf{L}_\tau \mathbf{U}\rangle_{\mathrm{F}} + \sum_{j\in\mathcal{L}} \ell(\mathbf{u}_j, \mathbf{y}_j) \\ &\to \frac{1}{2}\langle A, \mathbf{\Lambda}_\tau A\rangle_{\mathrm{F}} + \sum_{j\in\mathcal{L}} \ell(\mathbf{e}_j^{\mathrm{T}} VA, \mathbf{y}_j) =: \tilde{\mathcal{J}}_\ell(\mathbf{A};\mathbf{Y}). \end{aligned} \tag{6}$$

The vector $\boldsymbol{\alpha} = \mathbf{V}^{\mathrm{T}}\mathbf{u}$ (or the matrix $\mathbf{A} = \mathbf{V}^{\mathrm{T}}\mathbf{U}$) is the projection of $\mathbf{u}$ (or $\mathbf{U}$) onto the $M$ corresponding eigenvectors. Since the eigenvectors are orthonormal ($\mathbf{V}^{\mathrm{T}}\mathbf{V} = \mathbf{I} \in \mathbb{R}^{M\times M}$), $\boldsymbol{\alpha}$ is the vector of coefficients of the representation of $\mathbf{u}$ *restricted to span of the eigenbasis of the first M eigenvectors of* $\mathbf{L}$, $\mathrm{span}\{\mathbf{v}_1, \cdots, \mathbf{v}_M\}$. By restricting to this subspace, we not only speed up computation of approximate solutions for $\hat{\mathbf{u}}$ since the latent space is of (much) smaller dimension ($M \ll N$), but we also reduce the spatial complexity of the covariance matrix for the Model Change acquisition function calculations.

Recalling the discussion of Sect. 2.2, we can interpret the graph-based SSL models that minimize the objective functions of 5 and 6 as MAP estimators of corresponding Bayesian posterior distributions, now in terms of the variables $\boldsymbol{\alpha}$ and $A$. In Sects. 3.2, 3.3, and 3.4 we apply the Laplace Approximation to these non-Gaussian posterior distributions to approximate the covariance structure. This approximation of the covariance matrix will also allow us to efficiently estimate the properties of the look-ahead models of Sect. 2.3 for computing the Model Change acquisition function of 3.2.3.

We note that spectral truncation is commonly used for semi-supervised graph-based learning prior to the recent work on active learning [5, 8, 20]. Thus it is not surprising that it can be used to speed up the computation of other acquisition functions besides Model Change. For example, V-OPT [26] and Σ-OPT [37] graph-based acquisition functions can be approximated with spectral truncation since, in their original form, they require the inversion of the graph Laplacian matrix. In fact, for the larger datasets in our experiments (Sect. 4), we utilize this approximation when comparing against these acquisition functions for a more straightforward comparison of the benefits of our proposed Model Change acquisition function. See Appendix D for a further discussion of how spectral truncation can be applied to V-OPT and Σ-OPT acquisition functions.

## 3.2 Binary Model

We first derive the Model Change acquisition function on the binary model (5) for the GR, Logistic, and Probit models. The derivation for the binary model follows similarly to [41], except now done on this "spectral truncation" modification.

### 3.2.1 Laplace Approximation of the Binary Model

The Laplace approximation of the posterior distribution corresponding to (5), *with respect to the variable* $\boldsymbol{\alpha} \in \mathbb{R}^M$, gives

$$\begin{aligned}
\boldsymbol{\alpha}|\mathbf{y} &\sim \mathcal{N}(\hat{\boldsymbol{\alpha}}, \hat{\mathbf{C}}_{\hat{\boldsymbol{\alpha}}}), \qquad \hat{\boldsymbol{\alpha}} = \underset{\boldsymbol{\alpha}\in\mathbb{R}^M}{\arg\min}\ \tilde{J}_\ell(\boldsymbol{\alpha};\mathbf{y}),\\
\nabla_{\boldsymbol{\alpha}} \tilde{J}_\ell(\boldsymbol{\alpha};\mathbf{y}) &= \boldsymbol{\Lambda}_\tau \boldsymbol{\alpha} + \mathbf{V}^{\mathrm{T}} \sum_{j\in\mathcal{L}} F(\mathbf{e}_j^{\mathrm{T}} V\boldsymbol{\alpha}, y_j)\mathbf{e}_j,\\
\nabla^2_{\boldsymbol{\alpha}} \tilde{J}_\ell(\boldsymbol{\alpha};\mathbf{y}) &= \boldsymbol{\Lambda}_\tau + \mathbf{V}^{\mathrm{T}}\left(\sum_{j\in\mathcal{L}} F'(\mathbf{e}_j^{\mathrm{T}} V\boldsymbol{\alpha}, y_j)\mathbf{e}_j\mathbf{e}_j^{\mathrm{T}}\right)V,\\
\implies \hat{\mathbf{C}}_{\boldsymbol{\alpha}} &= \left(\nabla^2_{\boldsymbol{\alpha}} \tilde{J}_\ell(\boldsymbol{\alpha};\mathbf{y})\right)^{-1}\\
&= \left(\boldsymbol{\Lambda}_\tau + \mathbf{V}^{\mathrm{T}}\left(\sum_{j\in\mathcal{L}} F'(\mathbf{e}_j^{\mathrm{T}} V\boldsymbol{\alpha}, y_j)\mathbf{e}_j\mathbf{e}_j^{\mathrm{T}}\right)V\right)^{-1},
\end{aligned}$$

where $F'(x,y)$ is the second derivative of the loss function $\ell(x,y)$ *with respect to the first variable* following the notation of [23, 41]. That is, $F(x,y) := \frac{\partial \ell}{\partial x}(x,y)$ and $F'(x,y) := \frac{\partial^2 \ell}{\partial x^2}(x,y)$. $\hat{\mathbf{C}}_{\boldsymbol{\alpha}}$ is symmetric because $\boldsymbol{\Lambda}_\tau$ is diagonal and the objective function $\tilde{J}_\ell(\mathbf{u};\mathbf{y})$ is differentiable. Recall that the mean of this Gaussian distribution $\hat{\boldsymbol{\alpha}}$ is the MAP estimator of the true posterior, which corresponds to the coefficients of the desired graph-based SSL model's classifier $\hat{\mathbf{u}}$ in the eigenbasis represented by $\mathbf{V}$. This Gaussian distribution is now in a form in which one could apply adaptations of acquisition functions like MBR [56], V-Opt [26], and $\Sigma$-Opt [37] that were originally defined on Gaussian models (e.g., GR and HF), only now using other convex loss functions besides the squared-error loss.[4]

### 3.2.2 Look-Ahead Updates

With $\hat{\boldsymbol{\alpha}}$ denoting the *current* model's ($\tilde{J}_\ell(\boldsymbol{\alpha};\mathbf{y})$) MAP estimator, let $\hat{\boldsymbol{\alpha}}^{k,\hat{y}_k}$ denote the *look-ahead* model's ($\tilde{J}_\ell^{k,\hat{y}_k}(\boldsymbol{\alpha};\mathbf{y},\hat{y}_k)$) MAP estimator. While for most loss functions one cannot directly compute $\hat{\boldsymbol{\alpha}}^{k,\hat{y}_k}$ as a rank-one update from $\hat{\boldsymbol{\alpha}}$, we compute an approximation $\tilde{\boldsymbol{\alpha}}^{k,\hat{y}_k} \approx \hat{\boldsymbol{\alpha}}^{k,\hat{y}_k}$ by computing a single step of Newton's Method on the look-ahead objective $\tilde{J}^{k,\hat{y}_k}(\boldsymbol{\alpha};\mathbf{y},\hat{y}_k)$, *starting with the current MAP estimator* $\hat{\boldsymbol{\alpha}}$

$$
\begin{aligned}
\tilde{\boldsymbol{\alpha}}^{k,\hat{y}_k} &= \hat{\boldsymbol{\alpha}} - \left(\nabla^2_{\boldsymbol{\alpha}}\tilde{J}_\ell^{k,\hat{y}_k}\left(\hat{\boldsymbol{\alpha}};\mathbf{y},\hat{y}_k\right)\right)^{-1}\left(\nabla^2_{\boldsymbol{\alpha}}\tilde{J}_\ell^{k,\hat{y}_k}\left(\hat{\boldsymbol{\alpha}};\mathbf{y},\hat{y}_k\right)\right)\\
&= \hat{\boldsymbol{\alpha}} - \left(\hat{\mathbf{C}}_{\hat{\boldsymbol{\alpha}}}^{-1} + F'(\mathbf{v}_k^{\mathrm{T}}\hat{\boldsymbol{\alpha}},\hat{y}_k)\mathbf{v}^k(\mathbf{v}^k)^{\mathrm{T}}\right)^{-1}\left(0 + F(\mathbf{v}_k^{\mathrm{T}}\hat{\boldsymbol{\alpha}},\hat{y}_k)\mathbf{v}_k\right)\\
&= \hat{\boldsymbol{\alpha}} - \left(\hat{\mathbf{C}}_{\hat{\boldsymbol{\alpha}}} - \hat{\mathbf{C}}_{\hat{\boldsymbol{\alpha}}}\mathbf{v}_k\left(\frac{1}{F'(\mathbf{v}_k^{\mathrm{T}}\hat{\boldsymbol{\alpha}},\hat{y}_k)} + \mathbf{v}_k^{\mathrm{T}}\hat{\mathbf{C}}_{\hat{\boldsymbol{\alpha}}}\mathbf{v}_k\right)^{-1}(\mathbf{v}^k)^{\mathrm{T}}\hat{\mathbf{C}}_{\hat{\boldsymbol{\alpha}}}\right)F(\mathbf{v}_k^{\mathrm{T}}\hat{\boldsymbol{\alpha}},\hat{y}_k)\mathbf{v}_k.
\end{aligned}
$$

This gives

$$
\tilde{\boldsymbol{\alpha}}^{k,\hat{y}_k} = \hat{\boldsymbol{\alpha}} - \frac{F(\mathbf{v}_k^{\mathrm{T}}\hat{\boldsymbol{\alpha}},\hat{y}_k)}{1 + F'(\mathbf{v}_k^{\mathrm{T}}\boldsymbol{\alpha},\hat{y}_k)\mathbf{v}_k^{\mathrm{T}}\hat{\mathbf{C}}_{\hat{\boldsymbol{\alpha}}}\mathbf{v}_k}\hat{\mathbf{C}}_{\hat{\boldsymbol{\alpha}}}\mathbf{v}_k, \tag{7}
$$

where in the second to last line we have used (A1) and we have introduced $\mathbf{v}_k := (\mathbf{e}_k^{\mathrm{T}}\mathbf{V})^{\mathrm{T}}$, the $k^{th}$ row of $\mathbf{V}$ as a column vector in $\mathbb{R}^M$. Note that in the second line, $\hat{\boldsymbol{\alpha}}$ satisfies $\nabla\tilde{J}(\hat{\boldsymbol{\alpha}};\mathbf{y}) = \boldsymbol{\Lambda}_\tau\hat{\boldsymbol{\alpha}} + \sum_{j\in\mathcal{L}} F((\mathbf{v}^j)^{\mathrm{T}}\hat{\boldsymbol{\alpha}}, y_j)\mathbf{e}_j = \mathbf{0}$ since it is the minimizer of $\tilde{J}(\boldsymbol{\alpha};\mathbf{y})$.

### 3.2.3 Model Change (MC) Acquisition Function

We now define the acquisition function that we term "Model Change" (MC). This criterion quantifies how much the underlying graph-based SSL classifier changes as a result of adding a node $k \in \mathcal{U}$ and hypothesized label $\hat{y}_k$; that is, we measure how much the model classifier, $\hat{\mathbf{u}} = \mathbf{V}\hat{\boldsymbol{\alpha}}$, would change under the look-ahead model, $\hat{\mathbf{u}}^{k,\hat{y}_k} = \mathbf{V}\hat{\boldsymbol{\alpha}}^{k,\hat{y}_k}$. We approximate this model change via the update $\tilde{\boldsymbol{\alpha}}^{k,\hat{y}_k}$ of (7). Previous works indicate that calculating the approximate change in a model (classifier) from the addition of an index $k$ and associated pseudo-label $\hat{y}_k$ is an effective criterion for active learning [10, 30, 41]. The present work extends the MC method in [41] and resembles the Data-Based norm method of [30], wherein the authors essentially use this idea of quantifying the potential change in the model classifier as an acquisition function in the setting of kernel-based interpolation. While exact look-ahead calculations are possible in the non-graph-based setting of [30] and the smaller dataset settings presented in [41], the current work allows for a broader range of methods (Table 1) by approximating look-ahead calculations via (7).

[4] Recall that the Laplace approximation of a Gaussian distribution is itself, and so the Laplace approximation of the GR model's posterior distribution is itself.

In particular, our prior work [41] introduced the MC acquisition function *without* spectral truncation approximation for binary classification, showing superior performance to other methods. However, as is well-known for graph-Laplacian based methods, using the full spectrum for multiclass or larger datasets is computationally prohibitive. With this in mind, we focus on spectral truncation for the computational benefits while still accurately approximating the look-ahead updates due to the global clustering structure captured in the corresponding eigenvectors (see Sect. 3.1).

Employing (7), we propose the following MC acquisition function:

$$
\begin{aligned}
\mathcal{A}(k) &= \min_{\hat{y}_k \in \{\pm 1\}} \|\hat{\mathbf{u}}^{k,\hat{y}_k} - \hat{\mathbf{u}}\|_2 \approx \min_{\hat{y}_k \in \{\pm 1\}} \left\|\tilde{\mathbf{u}}^{k,\hat{y}_k} - \hat{\mathbf{u}}\right\|_2 = \min_{\hat{y}_k \in \{\pm 1\}} \left\|\tilde{\boldsymbol{\alpha}}^{k,\hat{y}_k} - \hat{\boldsymbol{\alpha}}\right\|_2 \\
&= \min_{\hat{y}_k \in \{\pm 1\}} \left| \frac{F(\mathbf{v}_k^{\mathrm{T}}\hat{\boldsymbol{\alpha}}, \hat{y}_k)}{1 + F'(\mathbf{v}_k^{\mathrm{T}}\hat{\boldsymbol{\alpha}}, \hat{y}_k)\mathbf{v}_k^{\mathrm{T}}\hat{\mathbf{C}}_{\hat{\boldsymbol{\alpha}}}\mathbf{v}_k} \right| \left\|\hat{\mathbf{C}}_{\hat{\boldsymbol{\alpha}}}\mathbf{v}_k\right\|_2 \\
&= \min_{\hat{y}_k \in \{\pm 1\}} \left| \frac{F(\hat{u}_k, \hat{y}_k)}{1 + F'(\hat{u}_k, \hat{y}_k)\mathbf{v}_k^{\mathrm{T}}\hat{\mathbf{C}}_{\hat{\boldsymbol{\alpha}}}\mathbf{v}_k} \right| \left\|\hat{\mathbf{C}}_{\hat{\boldsymbol{\alpha}}}\mathbf{v}_k\right\|_2,
\end{aligned}
$$

since the columns of $V$ are orthonormal and $\hat{u}_k = \mathbf{v}_k^{\mathrm{T}}\hat{\boldsymbol{\alpha}}$. Recall that $\hat{y}_k$ is the current "pseudo-label" for node $k$ as given by the current classifier $\hat{\mathbf{u}}$. We can write this acquisition function explicitly for each considered binary classification model as

$$
\begin{aligned}
\mathcal{A}_{\mathrm{GR}}(k) &= \frac{|\hat{u}_k - \mathrm{sgn}(\hat{u}_k)|}{\gamma^2 + \mathbf{v}_k^{\mathrm{T}}\hat{\mathbf{C}}_{\hat{\boldsymbol{\alpha}}}\mathbf{v}_k} \|\hat{\mathbf{C}}_{\hat{\boldsymbol{\alpha}}}\mathbf{v}_k\|_2, \\
\mathcal{A}_{\mathrm{L}}(k) &= \frac{\gamma(1 + \mathrm{e}^{|\hat{u}_k|/\gamma})}{\gamma^2(1 + \mathrm{e}^{|\hat{u}_k|/\gamma})^2 + \mathrm{e}^{|\hat{u}_k|/\gamma}\mathbf{v}_k^{\mathrm{T}}\hat{\mathbf{C}}_{\hat{\boldsymbol{\alpha}}}\mathbf{v}_k} \|\hat{\mathbf{C}}_{\hat{\boldsymbol{\alpha}}}\mathbf{v}_k\|_2, \\
\mathcal{A}_{\mathrm{P}}(k) &= \frac{\psi_\gamma(|\hat{u}_k|)\Psi_\gamma(|\hat{u}_k|)}{\left|\Psi_\gamma^2(|\hat{u}_k|) + \left(\psi_\gamma^2(|\hat{u}_k|) - \psi_\gamma'(|\hat{u}_k|)\Psi_\gamma(|\hat{u}_k|)\right)\mathbf{v}_k^{\mathrm{T}}\hat{\mathbf{C}}_{\hat{\boldsymbol{\alpha}}}\mathbf{v}_k\right|} \|\hat{\mathbf{C}}_{\hat{\boldsymbol{\alpha}}}\mathbf{v}_k\|_2,
\end{aligned}
$$

where $\hat{u}_k\hat{y}_k = \hat{u}_k\mathrm{sgn}(\hat{u}_k) = |\hat{u}_k|$. Note that in the GR model case, this notion of "Model Change" as calculated is exact; the value $\mathcal{A}_{\mathrm{GR}}(k)$ is exactly how much the underlying GR classifier would change by selecting $k$ with the label $\hat{y}_k = \mathrm{sgn}(\hat{u}_k)$.

### 3.3 Multiclass Gaussian Regression

We now apply the derivation of the acquisition function to the MGR case. Recalling (6), then the gradient and Hessian are

$$\nabla_A \tilde{\mathcal{J}}_{\rm GR}(\mathbf{A};\mathbf{Y}) = \mathbf{\Lambda}_\tau \mathbf{A} + \frac{1}{\gamma^2}\mathbf{V}^{\rm T}\mathbf{P}^{\rm T}(\mathbf{PVA}-\mathbf{Y}),$$
$$\nabla_A^2 \tilde{\mathcal{J}}_{\rm GR}(\mathbf{A};\mathbf{Y}) = \mathbf{\Lambda}_\tau + \mathbf{V}^{\rm T}\left(\frac{1}{\gamma^2}\mathbf{P}^{\rm T}\mathbf{P}\right)\mathbf{V},$$

which then give the Gaussian posterior distribution

$$\mathbf{A}|\mathbf{Y} \sim \mathcal{N}(\hat{\mathbf{A}}, \mathbf{I}_{n_c} \times \mathbf{C}_{\hat{\mathbf{A}}}), \quad \hat{\mathbf{A}} = \mathop{\arg\min}_{A\in\mathbb{R}^{M\times n_c}} \tilde{\mathcal{J}}_{\rm GR}(\mathbf{A};\mathbf{Y}),$$
$$\mathbf{C}_{\hat{\mathbf{A}}} = \left(\nabla_{\boldsymbol{\alpha}}^2 \tilde{\mathcal{J}}_{\rm GR}(\hat{\mathbf{A}};\mathbf{Y})\right)^{-1} = \left(\mathbf{\Lambda}_\tau + \mathbf{V}^{\rm T}\left(\frac{1}{\gamma^2}\mathbf{P}^{\rm T}\mathbf{P}\right)\mathbf{V}\right)^{-1}.$$

The posterior mean update (which is exact for MGR) becomes

$$\begin{aligned}\hat{\mathbf{A}}^{k,\hat{y}_k} &= \hat{\mathbf{A}} - \left(\nabla_A^2 \tilde{\mathcal{J}}_{\rm GR}^{k,\hat{y}_k}(\hat{\mathbf{A}};\mathbf{Y},\hat{\mathbf{y}}_k)\right)^{-1}\left(\nabla_A \tilde{\mathcal{J}}_{\rm GR}^{k,\hat{y}_k}(\hat{\mathbf{A}};\mathbf{Y},\hat{\mathbf{y}}_k)\right)\\ &= \hat{\mathbf{A}} - \frac{1}{\gamma^2+\mathbf{v}_k^{\rm T}\mathbf{C}_{\hat{\mathbf{A}}}\mathbf{v}_k}\mathbf{C}_{\hat{\mathbf{A}}}\mathbf{v}_k\left(\mathbf{v}_k^{\rm T}\hat{\mathbf{A}} - (\hat{\mathbf{y}}_k)^{\rm T}\right).\end{aligned} \tag{A1}$$

With the pseudolabel $\hat{y}_k = \mathop{\arg\max}_{c=1,2,\cdots,n_c} \hat{\mathbf{U}}_{k,c}$ and corresponding one-hot encoding $\hat{\mathbf{y}}_k \in \mathbb{R}^{n_c}$, the MC acquisition function becomes (using the relation $\mathbf{U} = \mathbf{VA}$)

$$\begin{aligned}\mathcal{A}_{\rm GR}(k) &= \left\|\hat{\mathbf{U}}^{k,\hat{y}_k} - \hat{\mathbf{U}}\right\|_{\rm F} = \left\|\hat{\mathbf{A}}^{k,\hat{y}_k} - \hat{\mathbf{A}}\right\|_{\rm F}\\ &= \left|\frac{1}{\gamma^2+\mathbf{v}_k^{\rm T}\mathbf{C}_{\hat{\mathbf{A}}}\mathbf{v}_k}\right|\left\|\mathbf{C}_{\hat{\mathbf{A}}}\mathbf{v}_k\left(\mathbf{v}_k^{\rm T}\hat{\mathbf{A}} - (\hat{\mathbf{y}}_k)^{\rm T}\right)\right\|_{\rm F}\\ &= \frac{1}{\gamma^2+\mathbf{v}_k^{\rm T}\mathbf{C}_{\hat{\mathbf{A}}}\mathbf{v}_k}\|\mathbf{C}_{\hat{\mathbf{A}}}\mathbf{v}_k\|_2\|\hat{\mathbf{A}}^{\rm T}\mathbf{v}_k - \hat{\mathbf{y}}_k\|_2,\end{aligned}$$

where we have used the orthonormality of the columns of $V$ and (A2).

## 3.4 Cross-Entropy (CE) Model

The softmax function in the CE model of Sect. 2.1.1 introduces dependencies between the columns of $\mathbf{U}$. For ease in calculations, let the vector $\mathbf{u} \in \mathbb{R}^{Nn_c}$ be the concatenation of the columns of $\mathbf{U}$. Likewise, define $\boldsymbol{\alpha} \in \mathbb{R}^{Mn_c}$ to be the concatenation of the columns of the matrix $A$. Further, define $\boldsymbol{\mathcal{V}} := {\rm diag}(\mathbf{V},\mathbf{V},\cdots,\mathbf{V}) \in \mathbb{R}^{Nn_c\times Mn_c}$ and $\mathbf{\Lambda}_\tau^{\otimes} := {\rm diag}\left(\mathbf{\Lambda}_\tau,\mathbf{\Lambda}_\tau,\cdots,\mathbf{\Lambda}_\tau\right) \in \mathbb{R}^{Mn_c\times Mn_c}$. We also define $\boldsymbol{\mathcal{P}}_j \in \mathbb{R}^{n_c\times Nn_c}$ to be the projection matrix that picks out the indices in $\mathbf{u} \in \mathbb{R}^{Nn_c}$ corresponding to node $j$; i.e., selecting the $j^{th}$ row of the related matrix $\mathbf{U} \in \mathbb{R}^{N\times n_c}$. Then $\mathbf{u} = \boldsymbol{\mathcal{V}}\boldsymbol{\alpha}$, and with $\hat{\mathbf{e}}_i$ denoting the $i^{th}$ standard basis vector in $\mathbb{R}^{Nn_c}$, the spectral truncation CE objective (3) can be written as

$$\begin{aligned}\tilde{\mathcal{J}}_{\rm CE}(\boldsymbol{\alpha};\mathbf{Y}) =& \frac{1}{2}\langle\boldsymbol{\alpha},\mathbf{\Lambda}_\tau^{\otimes}\boldsymbol{\alpha}\rangle\\ &+\sum_{j\in\mathcal{L}}\left(-\frac{1}{\gamma}\langle\mathbf{y}_j,\boldsymbol{\mathcal{P}}_j\boldsymbol{\mathcal{V}}\boldsymbol{\alpha}\rangle + \ln\left(\sum_{h=1}^{n_c} {\rm e}^{\langle\boldsymbol{\alpha},\boldsymbol{\mathcal{V}}^{\rm T}\hat{\mathbf{e}}_{j+(h-1)N}\rangle/\gamma}\right)\right),\end{aligned} \tag{8}$$

where $\mathbf{u} = \mathcal{V}\boldsymbol{\alpha}$ and $\mathcal{V}^{\mathrm{T}}\mathcal{V} = \mathbf{I}_{Mn_c}$ by orthonormality of the eigenvectors. Defining

$$\boldsymbol{\pi}^j_c = \frac{\mathrm{e}^{\langle \boldsymbol{\alpha}, \mathcal{V}^{\mathrm{T}}\hat{\mathbf{e}}_{j+(c-1)N}\rangle/\gamma}}{\sum_{h=1}^{n_c} \mathrm{e}^{\langle \boldsymbol{\alpha}, \mathcal{V}^{\mathrm{T}}\hat{\mathbf{e}}_{j+(h-1)N}\rangle/\gamma}} \quad \text{and} \quad \boldsymbol{\pi}^j := (\boldsymbol{\pi}^j_1, \cdots, \boldsymbol{\pi}^j_{n_c})^{\mathrm{T}} \in \mathbb{R}^{n_c},$$

the gradient and Hessian of (8) are

$$\nabla \tilde{\mathcal{J}}_{\mathrm{CE}}(\boldsymbol{\alpha}; \mathbf{Y}) = \boldsymbol{\Lambda}^{\otimes}_{\tau}\mathcal{V} + \frac{1}{\gamma}\mathcal{V}^{\mathrm{T}} \sum_{j\in\mathcal{L}} \mathcal{P}_j^{\mathrm{T}}\left(\boldsymbol{\pi}^j - \mathbf{y}_j\right),$$

$$\nabla^2 \tilde{\mathcal{J}}_{\mathrm{CE}}(\boldsymbol{\alpha}; \mathbf{Y}) = \boldsymbol{\Lambda}^{\otimes}_{\tau} + \frac{1}{\gamma^2}\mathcal{V}^{\mathrm{T}}\left(\mathbf{D}_{\mathcal{L}}(\boldsymbol{\alpha}) - \boldsymbol{\Pi}_{\mathcal{L}}(\boldsymbol{\alpha})\boldsymbol{\Pi}^{\mathrm{T}}_{\mathcal{L}}(\boldsymbol{\alpha})\right)\mathcal{V},$$

where we refer the reader to the supplementary material (Appendix B) for full calculation details. The Laplace approximation for the CE model yields

$$\boldsymbol{\alpha}|\mathbf{y} \sim \mathcal{N}(\hat{\boldsymbol{\alpha}}, \mathcal{C}_{\hat{\boldsymbol{\alpha}}}), \quad \hat{\boldsymbol{\alpha}} = \mathop{\arg\min}_{\boldsymbol{\alpha}\in\mathbb{R}^{Mn_c}} \tilde{\mathcal{J}}_{\mathrm{CE}}(\boldsymbol{\alpha}; \mathbf{Y}),$$

$$\mathcal{C}_{\hat{\boldsymbol{\alpha}}} = \left(\nabla^2 \tilde{\mathcal{J}}_{\mathrm{CE}}(\boldsymbol{\alpha}; \mathbf{Y})\right)^{-1} = \left(\boldsymbol{\Lambda}^{\otimes}_{\tau} + \frac{1}{\gamma^2}\mathcal{V}^{\mathrm{T}}\left(\mathbf{D}_{\mathcal{L}}(\boldsymbol{\alpha}) - \boldsymbol{\Pi}_{\mathcal{L}}(\boldsymbol{\alpha})\boldsymbol{\Pi}^{\mathrm{T}}_{\mathcal{L}}(\boldsymbol{\alpha})\right)\mathcal{V}\right)^{-1},$$

where we emphasize the dependence of $\mathcal{C}_{\boldsymbol{\alpha}}$ on the variable $\boldsymbol{\alpha}$, specifically taking the value $\mathcal{C}_{\hat{\boldsymbol{\alpha}}}$ at the MAP estimator $\hat{\boldsymbol{\alpha}}$. The inverse $\mathcal{C}_{\hat{\boldsymbol{\alpha}}} = \left(\nabla^2 \tilde{\mathcal{J}}(\hat{\boldsymbol{\alpha}})\right)^{-1} \in \mathbb{R}^{Mn_c \times Mn_c}$ is not prohibitively costly to compute because of its restricted size. Referring to the calculations in the supplementary material (Appendix B) the look-ahead calculations become

$$\nabla \tilde{\mathcal{J}}^{k,\hat{y}_k}_{\mathrm{CE}}(\hat{\boldsymbol{\alpha}}; \mathbf{Y}, \hat{\mathbf{y}}_k) = \nabla \tilde{\mathcal{J}}_{\mathrm{CE}}(\hat{\boldsymbol{\alpha}}; \mathbf{Y}) + \frac{1}{\gamma}\mathcal{V}_k^{\mathrm{T}}\left(\boldsymbol{\pi}^k - \hat{\mathbf{y}}_k\right), \quad \boldsymbol{\pi}^k = \left(\boldsymbol{\pi}^k_1, \cdots, \boldsymbol{\pi}^k_{n_c}\right)^{\mathrm{T}},$$

$$\nabla^2 \tilde{\mathcal{J}}^{k,\hat{y}_k}_{\mathrm{CE}}(\hat{\boldsymbol{\alpha}}; \mathbf{Y}, \hat{\mathbf{y}}_k) = \mathcal{C}^{-1}_{\hat{\boldsymbol{\alpha}}} + \frac{1}{\gamma^2}\mathcal{V}_k^{\mathrm{T}}\mathbf{B}_k\mathcal{V}_k,$$

where $\mathcal{V}_k := \mathcal{P}_k\mathcal{V}$ and $\mathbf{B}_k := \mathrm{diag}\left(\boldsymbol{\pi}^k\right) - \boldsymbol{\pi}^k\left(\boldsymbol{\pi}^k\right)^{\mathrm{T}}$. A Cholesky decomposition of the positive semi-definite $\mathbf{B}_k = \mathbf{T}_k^{\mathrm{T}}\mathbf{T}_k$ and using (A1) twice enable the approximation[5]

$$\begin{aligned}
\tilde{\boldsymbol{\alpha}}^{k,\hat{y}_k} &= \hat{\boldsymbol{\alpha}} - \left(\nabla^2 \tilde{\mathcal{J}}^{k,\hat{y}_k}_{\mathrm{CE}}(\hat{\boldsymbol{\alpha}}; \mathbf{Y}, \hat{\mathbf{y}}_k)\right)^{-1}\left(\nabla \tilde{\mathcal{J}}^{k,\hat{y}_k}_{\mathrm{CE}}(\hat{\boldsymbol{\alpha}}; \mathbf{Y}, \hat{\mathbf{y}}_k)\right) \\
&= \hat{\boldsymbol{\alpha}} - \frac{1}{\gamma}\mathcal{C}_{\hat{\boldsymbol{\alpha}}}\mathcal{V}_k^{\mathrm{T}}\left(\mathbf{I} - \mathbf{T}_k^{\mathrm{T}}\left(\mathbf{I} + \mathbf{T}_k\mathbf{G}_k\mathbf{T}_k^{\mathrm{T}}\right)^{-1}\mathbf{T}_k\mathbf{G}_k\right)\left(\boldsymbol{\pi}^k - \hat{\mathbf{y}}_k\right),
\end{aligned}$$

where $\mathbf{G}_k = \frac{1}{\gamma^2}\mathcal{V}_k\mathcal{C}_{\hat{\boldsymbol{\alpha}}}\mathcal{V}_k^{\mathrm{T}}$. With pseudo-label $\hat{y}_k$ and corresponding one-hot encoding $\hat{\mathbf{y}}_k$, the MC acquisition function for the CE spectral truncation modification becomes

$$\begin{aligned}
\mathcal{A}_{\mathrm{CE}}(k) &= \|\hat{\mathbf{u}} - \tilde{\mathbf{u}}^{k,\hat{y}_k}\|_2 = \|\hat{\boldsymbol{\alpha}} - \tilde{\boldsymbol{\alpha}}^{k,\hat{y}_k}\|_2 \\
&= \frac{1}{\gamma}\left\|\mathcal{C}_{\hat{\boldsymbol{\alpha}}}\mathcal{V}_k^{\mathrm{T}}\left(\mathbf{I} - \mathbf{T}_k^{\mathrm{T}}\left(\mathbf{I} + \mathbf{T}_k\mathbf{G}_k\mathbf{T}_k^{\mathrm{T}}\right)^{-1}\mathbf{T}_k\mathbf{G}_k\right)\left(\boldsymbol{\pi}^k - \hat{\mathbf{y}}_k\right)\right\|_2.
\end{aligned}$$

[5] Again, see Appendix B for more details.

This is efficient to compute because calculating $\mathbf{I} - \mathbf{T}_k^{\mathrm{T}}\left(\mathbf{I} + \mathbf{T}_k\mathbf{G}_k\mathbf{T}_k^{\mathrm{T}}\right)^{-1}\mathbf{T}_k\mathbf{G}_k \in \mathbb{R}^{n_c \times n_c}$ involves only $n_c \times n_c$ matrices.

## 4 Experiments & Numerics

This section presents empirical results for the Model Change (MC) acquisition function compared to other active learning acquisition functions in the various graph-based SSL models of Table 1. We reference each acquisition function in the form [*abbreviation of acquisition function*]-[*abbreviation of underlying model*]; for example **MC-GR** denotes the MC acquisition function in the Gaussian Regression (GR) model. The acquisition function acronyms are: **MC** (Model Change), **UNC** (Uncertainty [48]), **RAND** (Random), **VOPT** (V-Opt [29]), and **SOPT** (Σ-Opt [37]). The underlying models considered are **GR** (Gaussian Regression, Sect. 3.2.3), **MGR** (Multiclass Gaussian Regression, Sect. 3.3), **LOG** (Logistic, Sect. 3.2.3), **P** (Probit, Sect. 3.2.3), and **CE** (Cross-Entropy, Sect. 3.4), all in the spectral truncation form of Sect. 3.1.

We showcase the method on a synthetic dataset (Binary-Clusters) as well as three real-world datasets: MNIST [34] and two hyperspectral imagery (HSI) datasets, Salinas A and Urban. On the binary tests, we include comparisons with the *data-based norm* acquisition function (**DB-RKHS**) [30] as well as the original V-Opt [29] and Σ-Opt [37] methods in the HF model [55], annotated as **VOPT-HF** and **SOPT-HF** in the plots below. We calculate the average accuracies over five trials *according to the underlying SSL classifier of the acquisition function*; that is, for the choices from the **MC-GR** acquisition function, we report the accuracies *in the GR model*. We straightforwardly adapt V-Opt and Σ-Opt methods for the MGR model to allow for comparison on the multiclass tests, see Appendix D.

All but one of the tests run in the "batch" mode of active learning, wherein for simplicity we select $B = 5$ points for the query set $\mathcal{Q}$ per active learning iteration. Per the discussion of Sect. 2.5, at each iteration the *candidate set* $\mathcal{S} \subset \mathcal{U}$ contains 10% of the total points in $\mathcal{U}$, sampled uniformly at random. The query set comprises the top $B = 5$ *maximizers* of the active learning acquisition function on $\mathcal{S}$. While an interesting question, this paper does not explore varying the batch size nor the candidate set size; we leave this for future work.

### 4.1 Graph Construction Settings

We construct similarity graphs using shared parameters across the different datasets. For the non-hyperspectral datasets (Binary-Clusters and MNIST), the similarity graph contains 10-nearest neighbors with weights $w_{ij}$ given by the Gaussian similarity kernel $\kappa(\mathbf{x}_i, \mathbf{x}_j) = \exp(\|\mathbf{x}_i - \mathbf{x}_j\|_2^2/\sigma)$, with $\sigma = 3$. For the hyperspectral datasets (Salinas A and Urban), the similarity graph is constructed using 15-nearest neighbors with weights $w_{ij}$ given by the cosine similarity $\kappa(\mathbf{x}_i, \mathbf{x}_j) = \langle \mathbf{x}_i, \mathbf{x}_j \rangle / (\|\mathbf{x}_i\|_2 \|\mathbf{x}_j\|_2)$. As is common in similarity graph construction, we employ Zelnik-Perona scaling [54], but only for the *non-hyperspectral* datasets. Due to the sparse nature of these similarity graphs, the $M = 50$ lowest eigenvalues of the graph's normalized Laplacian matrix are calculated with standard sparse eigensolvers.[6] In the binary experiments, the eigenvalue perturbation $\tau$ (Sect. 3) is set to $\tau = 0.001$, while $\tau = 0.005$ for the multiclass experiments. For the binary experiments

[6] One could use the Nyström extension [4] for calculating approximate eigenvalues and eigenvectors in the case that constructing such a k-nearest neighbors similarity graph is prohibitively costly [5].

(**GR, LOG,** and **P**), the loss parameter is set to $\gamma = 0.5$. We use the reported settings of $h = 0.1$ in the **RKHS** model [30] and $\delta = 0.01$ in the **HF** model [56] in the corresponding experiments. For the multiclass experiments, $\gamma = 0.1, 0.5$ in the **MGR** and **CE** models, respectively.

### 4.2 Binary-Clusters

Binary-Clusters is a synthetically created dataset we created consisting of various clusters of differing sizes, locations, and spreads. The ground-truth classification of these clusters is shown in each of the plots of Sect. 4 as either red or blue dots. In our code[7] we provide the code for recreating this particular dataset with a function entitled create_binary_clusters(). We run two tests: one that selects 100 query points *sequentially* ($B = 1$ per iteration) and the other that selects 100 query points in *batches* ($B = 5$ per iteration). While one could consider designing a criterion for terminating the active learning process, this is outside the scope of the present work.[8] As such, our tests are run for a fixed number of iterations. Both tests begin with only 2 initially labeled points, one in each class. Successful active learning on this dataset requires "exploring" the many different regions (squares) as well as "exploiting" the true decision boundaries between adjacent squares efficiently while the underlying classifier improves.

### 4.3 MNIST

The MNIST [34] dataset contains 70 000 grayscale $28 \times 28$ pixel images of handwritten digits (0–9). We represent each image as a 784-dimensional vector $\mathbf{x}_i$ and normalize the pixel values to range from 0 to 1. Each trial begins with 20 labeled points (i.e., $\approx 0.03\%$ of the total, 2 points per class) and then selects 500 active learning points in batches of size $B = 5$. We consider the graph containing the full 70 000 points in the MNIST dataset and calculate the accuracy over the unlabeled set at each iteration, not a held-out "testing set". This is more relevant to the SSL framework, as opposed to supervised learning. In panel (a) of Figs. 4 and 5 we plot the results for the various acquisition functions applied in the **MGR** and **CE** models, respectively.

### 4.4 Salinas A Hyperspectral Imagery Dataset

The Salinas A dataset is a common Hyperspectral Imagery (HSI) dataset containing 7 138 total pixels in a $83 \times 86$ image in $d = 224$ wavelengths. This is an image of Salinas, USA taken with the Aviris sensor; this image contains 6 classes of plant types arranged in a diagonal pattern (see Fig. 6a). The goal is to classify the pixels $\mathbf{x}_i \in \mathbb{R}^d$ in the image into material classes based on the samples from the $d$ different wavelengths. The "ground-truth" classification is shown in Fig. 6a. Each trial begins with two initially labeled points per class and then selects 500 active learning points in batches of size $B = 5$. In panel (b) of Figs. 4 and 5 we plot the resulting accuracy curves for the various acquisition functions on this task evaluated in the **MGR** and **CE** models, respectively.

[7] https://github.com/millerk22/model-change-paper

[8] In related work, the authors in [13] actually used the distribution of the current values of our Model Change acquisition function as a terminating condition.

### 4.5 Urban Hyperspectral Imagery Dataset

The Urban dataset is another common HSI dataset which contains 94 249 total pixels in a $307 \times 307$ image where each pixel represents a $2 \times 2$ m$^2$ area. The original hyperspectral image contains 210 wavelengths sampled ranging from 400 nm to 2 500 nm, resulting in a spectral resolution of 10 nm. As is typical in HSI, atmospheric effects and dense water vapor contaminate a number of the wavelengths and so we consider only the remaining 162 wavelengths. Pixels belong to one of six categories: *asphalt*, *grass*, *tree*, *roof*, *metal*, and *dirt*. Figure 6b shows the "ground-truth" classification. Each trial begins with two initially labeled points per class and then selects 500 active learning points in batches of size $B = 5$. In panel (c) of Figs. 4 and 5 we plot the resulting accuracy curves for the various acquisition functions on this task evaluated in the **MGR** and **CE** models, respectively.

### 4.6 Discussion of Method Performance and Comparison

For an acquisition function to be useful in the active learning process, its accuracy curve should both (i) initially increase rapidly compared to other methods and (ii) not level off at lower final accuracy. In the binary classification tests, i.e., the sequential and batch tests on the Binary-Clusters dataset (Figs. 3a and b), the **VOPT** and **SOPT** acquisition functions in the **HF** model perform well early on, but level off at a lower overall accuracy. By comparison, the **MC** accuracy curves increase slightly slower in the beginning, but achieve a higher overall accuracy than **VOPT** and **SOPT**. The **DB-RKHS** method performs very well in both tests, but we note this model and associated acquisition function are more costly to compute than our spectral-truncation models, see Fig. 7. All methods perform better than uncertainty sampling (**UNC-LOG**[9]) in both tests, which is especially slow to increase the accuracy early on in the sequential test.

Section 4 shows the distribution of active learning choices for a few of the considered methods in the binary sequential test, allowing us a glimpse at the empirical characteristics of each acquisition function's choices. Note that the **VOPT-HF** choices (Fig. 2b) are nearly spread out evenly among the whole unit square domain, while the **MC-GR** (Fig. 2c), and **DB-RKHS** (Fig. 2d) choices include points in every square, but also contain a higher concentration of choices along the boundaries between the squares. The **MC-GR** and **DB-RKHS** methods not only *explore the extent of the domain of the dataset*, but also *exploit classification information* by selecting points along the boundaries between the clusters throughout the whole domain. In contrast, the **VOPT-HF** method selects points evenly spread out over the whole domain, which arguably helps this method to achieve a beneficial increase early on in the active learning process, but does not transition to exploiting known classification information along the decision boundaries. **UNC-LOG** (Fig. 2a) in this run chooses points that lie between the long, tall blue cluster on the right and the top-right red cluster, while ignoring various other clusters in the dataset; in a sense, **UNC-LOG** *exploits* the known classification information without sufficiently exploring the extent of the dataset's domain.

[9] We report only **UNC-LOG** as performs best of all the three models **GR**, **LOG**, and **P**.

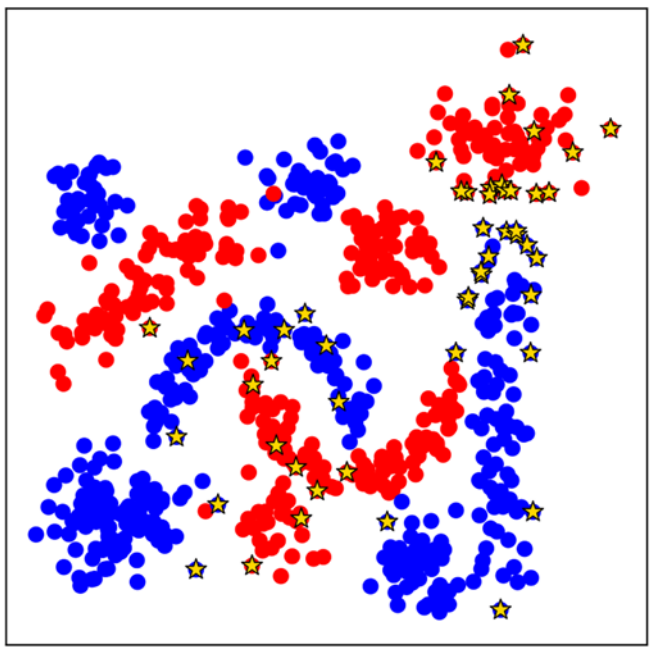

(a) **UNC-LOG**

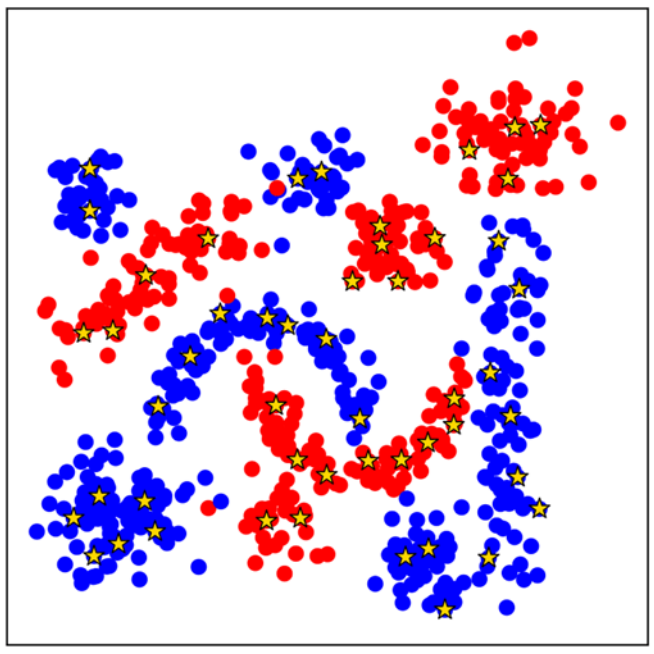

(b) **VOPT-HF**

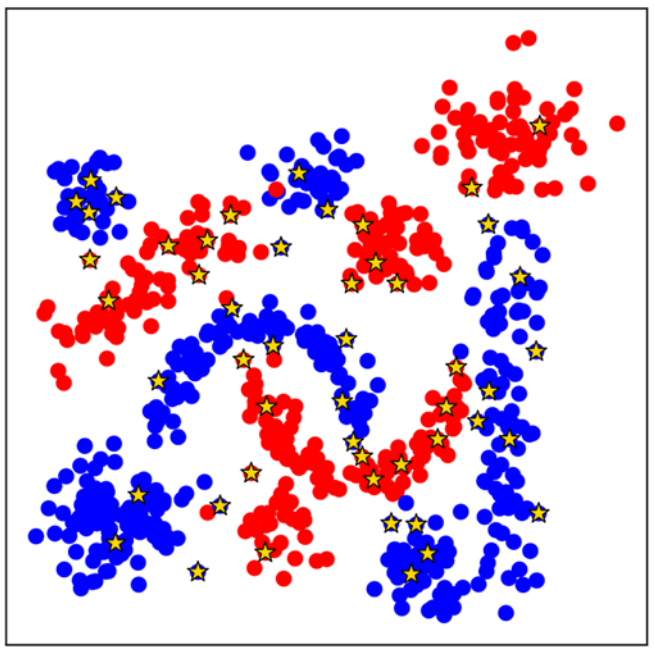

(c) **MC-GR**

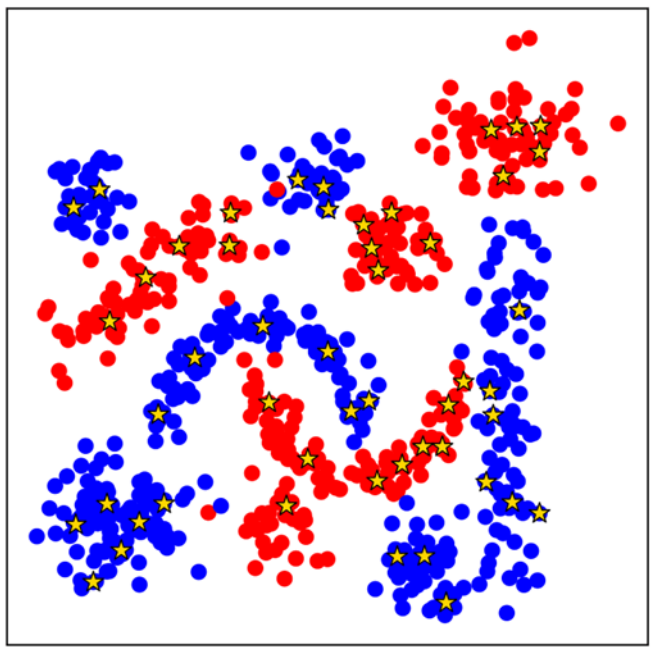

(d) **DB-RKHS**

**Fig. 2** Binary-Clusters Sequential Active Learning Choices. Each plot shows the ground truth classification (red/blue) for Binary-Clusters dataset as well as the active learning choices (yellow stars) for an assortment of acquisition functions. **UNC-LOG** selects points that are between only some clusters, while **VOPT-HF** selects points evenly spread out over the whole domain. In either case, overall performance is eventually suboptimal compared to **DB-RKHS**, and **MC-GR** methods which select points located in each of the clusters as well as along the boundary between oppositely labeled clusters

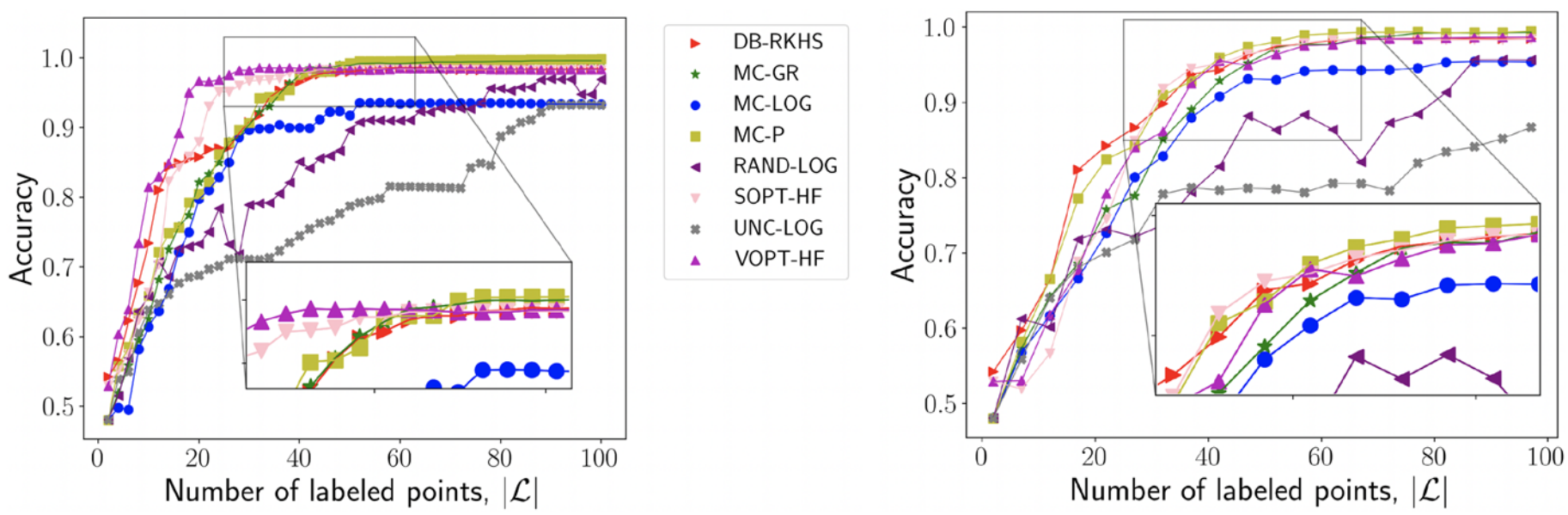

(a) Sequential, $B = 1$ (b) Batch, $B = 5$

**Fig. 3** Binary-Clusters accuracy plots, with 2 initially labeled points in each experiment. Sequential (a) performs 200 active learning iterations selecting $B = 1$ point per iteration, while Batch (b) performs 100 active learning iterations selecting $B = 5$ points per iteration. **VOPT-HF** and **SOPT-HF** achieve a quick initial accuracy increase in the sequential test (a), but level off at a lower accuracy than **MC-GR**, **MC-P**, and **DB-RKHS**; the **HF** and **RKHS** acquisition functions are however more costly to compute (Fig. 7). Zoomed-in pane in panel (a) highlights the transition point when **MC-GR** and **MC-P** achieve higher overall accuracy compared to **VOPT-HF** and **SOPT-HF** whose accuracy has leveled off

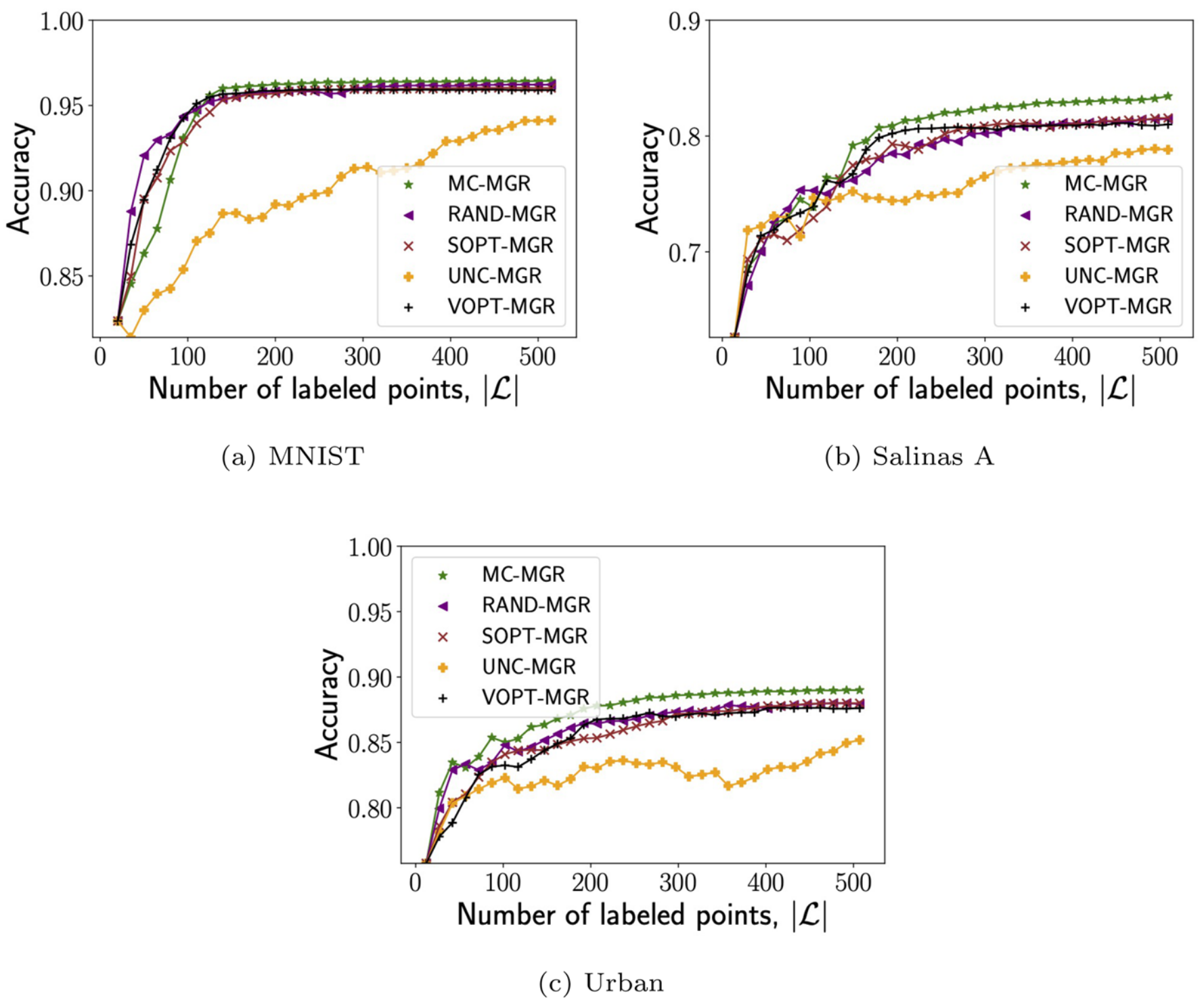

**Fig. 4** Accuracy plots for acquisition functions in the **MGR** model applied to MNIST, Salinas A, and Urban datasets. For each dataset, two points are selected uniformly at random from each class to initially label and then acquisition functions select 500 points in 100 batches of size $B = 5$

While the **DB-RKHS** method is very similar in flavor to our MC acquisition function, it is more computationally expensive both in model training as well as in acquisition function evaluation (see Fig. 7) because of dense similarity kernel computations[10]. Likewise, the **HF** model requires a matrix inversion of a large submatrix of the graph Laplacian which is undesirable for scaling to larger problems. By restricting our underlying classifier to the span of only a subset of the eigenvalues and eigenvectors of the graph Laplacian $L$, we achieve faster model training and acquisition function evaluation. Despite this significant model compression the present work is competitive with these more costly methods and models.

For all the multiclass tests (Sect. 4.3), the **MC-CE** acquisition function performs the best at both increasing the underlying model's accuracy early on and obtaining the highest accuracy overall in the active learning process. The **CE** classifier initially has *lower* accuracy than the **MGR** classifier in the MNIST and Urban tests, but quickly achieves a higher accuracy. While one can remedy the low initial accuracy of the **CE** model in practice by hyperparameter tuning, we set the hyperparameters to be consistent across the shown datasets so as to showcase the efficacy of the acquisition function

[10] One could approximate the kernel (e.g., Nyström extension), but we report the original formulation of [30].

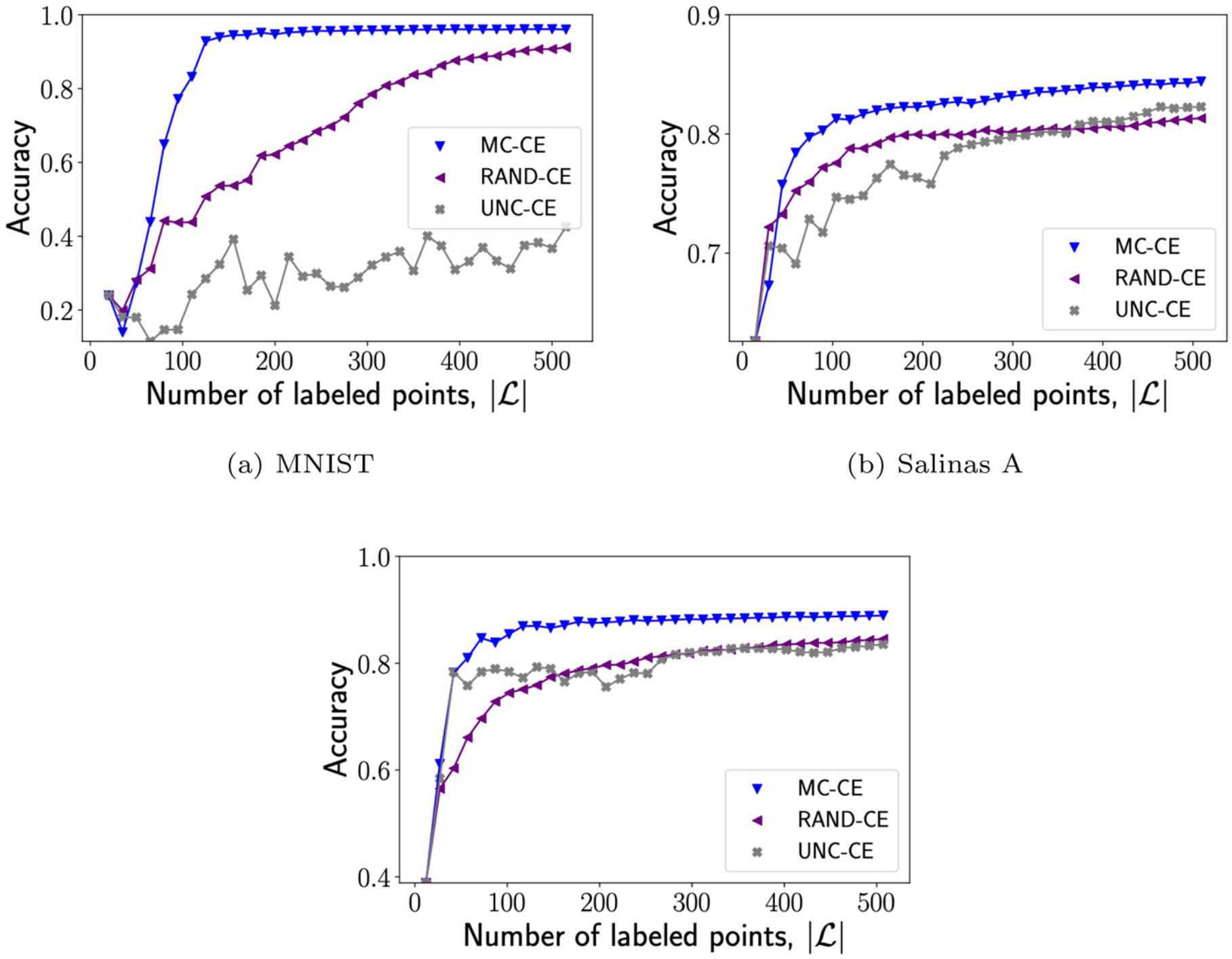

(a) MNIST (b) Salinas A

(c) Urban

**Fig. 5** Accuracy plots for acquisition functions in the **CE** model applied to MNIST, Salinas A, and Urban datasets. For each dataset, two points are selected uniformly at random from each class to initially label and then acquisition functions select 500 points in 100 batches of size $B = 5$

regardless of hyperparameter tuning. Further, the aim of the active learning process is to iteratively choose subsets of points to improve the underlying classifier and ultimately achieve the highest accuracy under the chosen model, not the design of the optimal underlying classifier in the presence of few labeled data.

We conclude this discussion with a note about the scalability of the acquisition function evaluations. Figure 7 shows the average time per active learning iteration to calculate the acquisition function on the candidate set, for increasing dataset size, $N$. Solid lines present timing results for binary models, while dashed lines present timing results for multiclass models. We exclude the **SOPT** results as they are indistinguishable from the **VOPT** results.

The family of binary **VOPT** and **MC** acquisition functions that utilize the spectral truncation approximation scale similarly, though the **P** (Probit) model's **MC** acquisition function has a significant overhead cost due to the repeated PDF and CDF calculations required for evaluating $F$ and $F'$ (see Sect. 3.2). In contrast, the **DB-RKHS** and **VOPT-HF** acquisition functions scale noticeably worse. This is to be expected since the spectral truncation applied to the other acquisition functions directly addresses the poor scaling of the **DB-RKHS** and **VOPT-HF** methods.

The multiclass acquisition functions (which all utilize the spectral truncation approximation) scale similarly to each other, though the **MC-CE** method has a greater

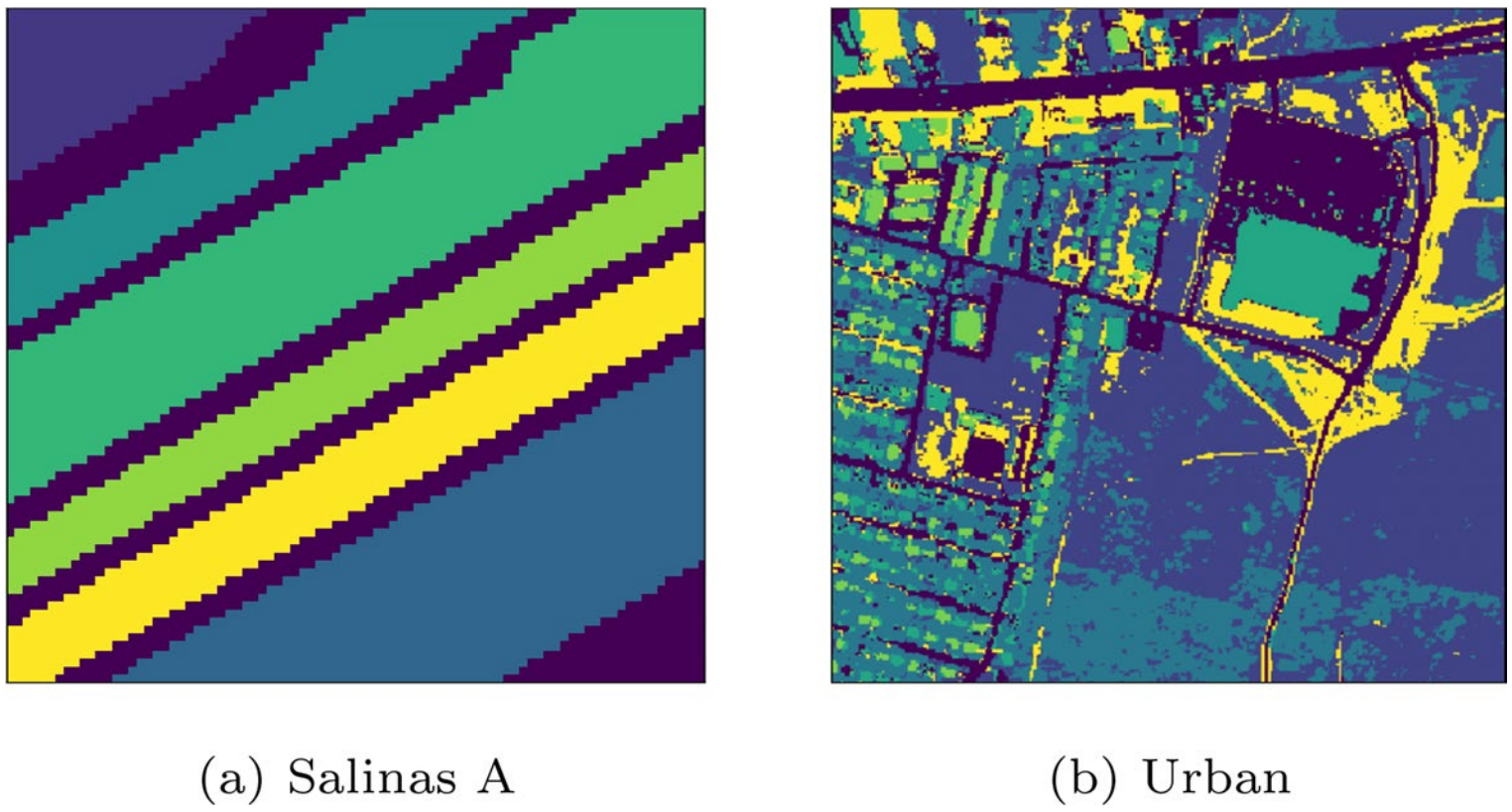

**Fig. 6** Hyperspectral Dataset Ground-Truth Classifications

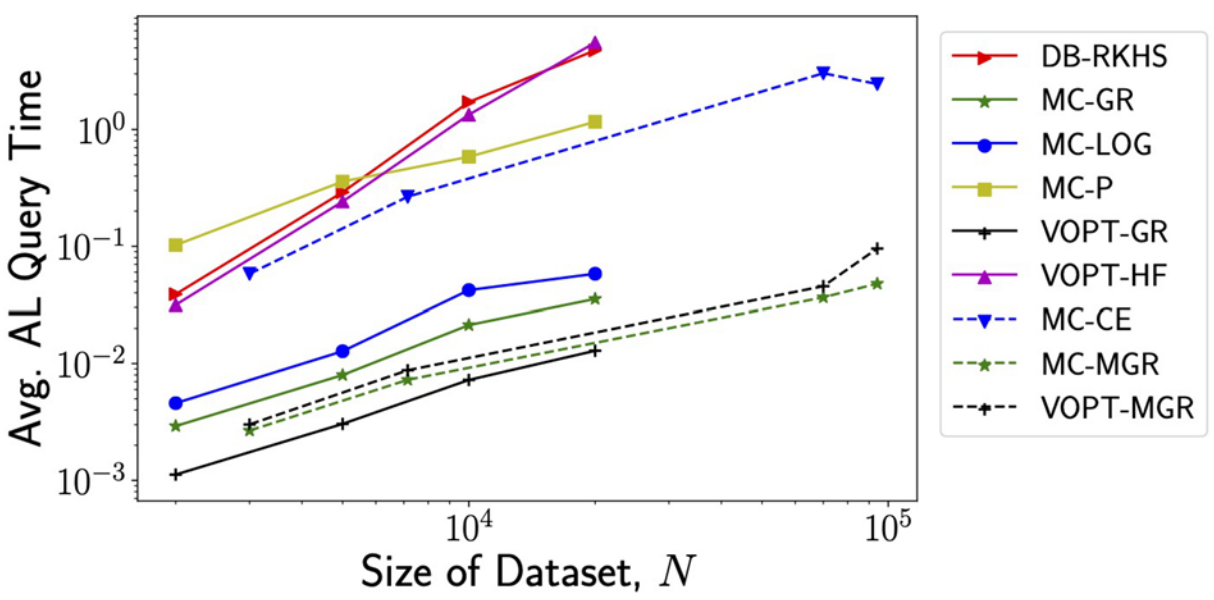

**Fig. 7** Average Active Learning Query Iteration Timing Comparison. Each iteration of the active learning process selects uniformly at random a candidate set comprising 10% of the unlabeled dataset on which to evaluate the corresponding acquisition function. Times shown are the averaged recorded time to evaluate the acquisition function on the candidate set for increasing overall dataset size, $N$. Solid lines present timing results for binary models, while dashed linear present timing results for multiclass models

overhead cost. The size of the posterior covariance matrix (i.e., inverse Hessian used in look-ahead calculations) for the **MGR** model ($\mathbf{C}_{\hat{\mathbf{A}}} \in \mathbb{R}^{M \times M}$) compared to the **CE** model ($\mathcal{C}_{\hat{\boldsymbol{\alpha}}} \in \mathbb{R}^{Mn_c \times Mn_c}$) straightforwardly clarifies this disparity. We conclude that the MC acquisition function adapted to the spectral truncation form of graph-based SSL provides both a scalable and effective method for active learning.

## 5 Conclusion

We present a novel Model Change (MC) active learning acquisition function along with a general framework that unifies different graph-based SSL models. Applying the Laplace approximation to the non-Gaussian Bayesian posterior distributions arising from different loss functions in the family of graph-based SSL models of Table 1 admits efficient approximations of how the underlying classifier could change as a result of labeling unlabeled points. This framework and associated active learning acquisition function are made more efficient by introducing the "spectral truncation" modifications, wherein we use only the lower-lying eigenvalues and corresponding eigenvectors in constraining the graph-based SSL classifiers as well as diminishing the memory requirements of the model. The MC acquisition function shows to be efficient for active learning compared to other methods natural for the graph-based SSL setting. Interesting future directions include (i) providing theoretical guarantees on the exploration versus exploitation properties of this MC acquisition function and (ii) adapting this MC framework to more interesting and complicated graph-based models.

## Appendix A Useful Identities

We briefly state two identities here that we use various times in the derivations hereafter.

**Theorem A1** (Woodbury Matrix Identity) *For matrices* $A \in \mathbb{R}^{p\times p}, C \in \mathbb{R}^{q\times q}, \mathbf{U} \in \mathbb{R}^{p\times q}$, $V \in \mathbb{R}^{q\times p}$ *with* $A$, $C$ *both invertible*, *then*

$$(\mathbf{A}+\mathbf{UCV})^{-1} = \mathbf{A}^{-1} - \mathbf{A}^{-1}\mathbf{U}\big(\mathbf{C}^{-1}+\mathbf{VA}^{-1}\mathbf{U}\big)^{-1}\mathbf{VA}^{-1}. \tag{A1}$$

**Theorem A2** *For vectors* $\mathbf{s} \in \mathbb{R}^p, \mathbf{t} \in \mathbb{R}^q$, *the Frobenius norm of the rank-one matrix* $\mathbf{st}^{\mathrm{T}} \in \mathbb{R}^{p\times q}$ *equals the product of the* 2-*norms of the individual vectors*:

$$\|\mathbf{st}^{\mathrm{T}}\|_{\mathrm{F}} = \sqrt{\mathrm{Tr}(\mathbf{ts}^{\mathrm{T}}\mathbf{st}^{\mathrm{T}})} = \|\mathbf{s}\|_2\sqrt{\mathrm{Tr}(\mathbf{tt}^{\mathrm{T}})} = \|\mathbf{s}\|_2\sqrt{\mathrm{Tr}(\mathbf{t}^{\mathrm{T}}\mathbf{t})} = \|\mathbf{s}\|_2\|\mathbf{t}\|_2. \tag{A2}$$

## Appendix B Gradient and Hessian Calculations for Cross-Entropy (CE) Model

Here we present the details of the gradient and Hessian calculations for the spectral truncation CE Model of Sect. 3.4. First, we compute the gradient of of $\tilde{\mathcal{J}}_{\mathrm{CE}}(\boldsymbol{\alpha};\mathbf{Y})$:

$$\begin{aligned}\nabla\tilde{\mathcal{J}}_{\mathrm{CE}}(\boldsymbol{\alpha};\mathbf{Y}) =& \boldsymbol{\Lambda}_\tau^{\otimes}\boldsymbol{\alpha} + \sum_{j\in\mathcal{L}}\frac{1}{\gamma}\left\{-\boldsymbol{\mathcal{V}}^{\mathrm{T}}\boldsymbol{\mathcal{P}}_j^{\mathrm{T}}\mathbf{y}^j + \frac{\sum_{c=1}^{n_c}\mathrm{e}^{\langle\boldsymbol{\alpha},\boldsymbol{\mathcal{V}}^{\mathrm{T}}\hat{\mathbf{e}}_{j+(c-1)N}\rangle/\gamma}\boldsymbol{\mathcal{V}}^{\mathrm{T}}\hat{\mathbf{e}}_{j+(c-1)N}}{\sum_{h=1}^{n_c}\mathrm{e}^{\langle\boldsymbol{\alpha},\boldsymbol{\mathcal{V}}^{\mathrm{T}}\hat{\mathbf{e}}_{j+(h-1)N}\rangle/\gamma}}\right\}\\ =& \boldsymbol{\Lambda}_\tau^{\otimes}\boldsymbol{\alpha} + \boldsymbol{\mathcal{V}}^{\mathrm{T}}\sum_{j\in\mathcal{L}}\frac{1}{\gamma}\left\{-\boldsymbol{\mathcal{P}}_j^{\mathrm{T}}\mathbf{y}^j + \sum_{c=1}^{n_c}\underbrace{\frac{\mathrm{e}^{\langle\boldsymbol{\alpha},\boldsymbol{\mathcal{V}}^{\mathrm{T}}\hat{\mathbf{e}}_{j+(c-1)N}\rangle/\gamma}}{\sum_{h=1}^{n_c}\mathrm{e}^{\langle\boldsymbol{\alpha},\boldsymbol{\mathcal{V}}^{\mathrm{T}}\hat{\mathbf{e}}_{j+(h-1)N}\rangle/\gamma}}}_{\boldsymbol{\pi}_c^j(\boldsymbol{\alpha})}\hat{\mathbf{e}}_{j+(c-1)N}\right\}\end{aligned}$$

and then we further compute

$$\begin{aligned}
\nabla \boldsymbol{\pi}_c^j(\boldsymbol{\alpha}) &= \frac{\nabla\left(\mathrm{e}^{\langle \boldsymbol{\alpha}, \boldsymbol{\mathcal{V}}^{\mathrm{T}}\hat{\mathbf{e}}_{j+(c-1)N}\rangle/\gamma}\right)}{\sum_{h=1}^{n_c}\mathrm{e}^{\langle \boldsymbol{\alpha}, \boldsymbol{\mathcal{V}}^{\mathrm{T}}\hat{\mathbf{e}}_{j+(h-1)N}\rangle/\gamma}} \\
&\quad - \frac{\mathrm{e}^{\langle \boldsymbol{\alpha}, \boldsymbol{\mathcal{V}}^{\mathrm{T}}\hat{\mathbf{e}}_{j+(h-1)N}\rangle/\gamma}}{\left(\sum_{h=1}^{n_c}\mathrm{e}^{\langle \boldsymbol{\alpha}, \boldsymbol{\mathcal{V}}^{\mathrm{T}}\hat{\mathbf{e}}_{j+(h-1)N}\rangle/\gamma}\right)^2}\nabla\left(\sum_{h=1}^{n_c}\mathrm{e}^{\langle \boldsymbol{\alpha}, \boldsymbol{\mathcal{V}}^{\mathrm{T}}\hat{\mathbf{e}}_{j+(h-1)N}\rangle/\gamma}\right) \\
&= \frac{\mathrm{e}^{\langle \boldsymbol{\alpha}, \boldsymbol{\mathcal{V}}^{\mathrm{T}}\hat{\mathbf{e}}_{j+(c-1)N}\rangle/\gamma}}{\sum_{h=1}^{n_c}\mathrm{e}^{\langle \boldsymbol{\alpha}, \boldsymbol{\mathcal{V}}^{\mathrm{T}}\hat{\mathbf{e}}_{j+(h-1)N}\rangle/\gamma}}\boldsymbol{\mathcal{V}}^{\mathrm{T}}\hat{\mathbf{e}}_{j+(c-1)N} \\
&\quad - \boldsymbol{\pi}_c^j\sum_{h=1}^{n_c}\frac{\mathrm{e}^{\langle \boldsymbol{\alpha}, \boldsymbol{\mathcal{V}}^{\mathrm{T}}\hat{\mathbf{e}}_{j+(h-1)N}\rangle/\gamma}}{\sum_{m=1}^{n_c}\mathrm{e}^{\langle \boldsymbol{\alpha}, \boldsymbol{\mathcal{V}}^{\mathrm{T}}\hat{\mathbf{e}}_{j+(m-1)N}\rangle/\gamma}}\boldsymbol{\mathcal{V}}^{\mathrm{T}}\hat{\mathbf{e}}_{j+(h-1)N} \\
&= \frac{\boldsymbol{\pi}_c^j(\boldsymbol{\alpha})}{\gamma}\boldsymbol{\mathcal{V}}^{\mathrm{T}}\left(\hat{\mathbf{e}}_{j+(c-1)N} - \sum_{h=1}^{n_c}\boldsymbol{\pi}_h^j(\boldsymbol{\alpha})\hat{\mathbf{e}}_{j+(h-1)N}\right).
\end{aligned}$$

With $\mathbf{e}_j \in \mathbb{R}^N$, we define

$$\mathbf{D}_{\mathcal{L}}(\boldsymbol{\pi}_c(\boldsymbol{\alpha})) = \sum_{j\in\mathcal{L}}\boldsymbol{\pi}_c^j(\boldsymbol{\alpha})\mathbf{e}_j\mathbf{e}_j^{\mathrm{T}}, \quad \boldsymbol{\Pi}_{\mathcal{L}}(\boldsymbol{\alpha}) = \begin{pmatrix}\mathbf{D}_{\mathcal{L}}(\boldsymbol{\pi}_1(\boldsymbol{\alpha}))\\ \mathbf{D}_{\mathcal{L}}(\boldsymbol{\pi}_2(\boldsymbol{\alpha}))\\ \vdots \\ \mathbf{D}_{\mathcal{L}}(\boldsymbol{\pi}_{n_c}(\boldsymbol{\alpha}))\end{pmatrix},$$

$$\mathbf{D}_{\mathcal{L}}(\boldsymbol{\alpha}) = \begin{pmatrix}\mathbf{D}_{\mathcal{L}}(\boldsymbol{\pi}_1(\boldsymbol{\alpha})) & 0 & \cdots & 0\\ 0 & \mathbf{D}_{\mathcal{L}}(\boldsymbol{\pi}_2(\boldsymbol{\alpha})) & \cdots & \vdots\\ \vdots & \vdots & & \vdots\\ 0 & \cdots & \cdots & \mathbf{D}_{\mathcal{L}}(\boldsymbol{\pi}_{n_c}(\boldsymbol{\alpha}))\end{pmatrix},$$

so that we can compute the Hessian

$$\begin{aligned}
\nabla^2\tilde{\mathcal{J}}_{\mathrm{CE}}(\boldsymbol{\alpha};\mathbf{Y}) &= \boldsymbol{\Lambda}_\tau^{\otimes} \\
&\quad + \frac{1}{\gamma^2}\boldsymbol{\mathcal{V}}^{\mathrm{T}}\left[\sum_{j\in\mathcal{L}}\sum_{c,h=1}^{n_c}\left(\boldsymbol{\pi}_c^j(\boldsymbol{\alpha})\delta_{ch} - \boldsymbol{\pi}_c^j(\boldsymbol{\alpha})\boldsymbol{\pi}_h^j(\boldsymbol{\alpha})\right)\hat{\mathbf{e}}_{j+(c-1)N}\hat{\mathbf{e}}_{j+(h-1)N}^{\mathrm{T}}\right]\boldsymbol{\mathcal{V}} \\
&= \boldsymbol{\Lambda}_\tau^{\otimes} + \frac{1}{\gamma^2}\boldsymbol{\mathcal{V}}^{\mathrm{T}}\left(\mathbf{D}_{\mathcal{L}}(\boldsymbol{\alpha}) - \boldsymbol{\Pi}_{\mathcal{L}}(\boldsymbol{\alpha})\boldsymbol{\Pi}_{\mathcal{L}}^{\mathrm{T}}(\boldsymbol{\alpha})\right)\boldsymbol{\mathcal{V}}.
\end{aligned}$$

Now turning to the look-ahead objective, we likewise introduce

$$\begin{aligned}
\mathbf{D}_{\{k\}}(\boldsymbol{\pi}_c(\boldsymbol{\alpha})) &= \boldsymbol{\pi}_c^k(\boldsymbol{\alpha})\mathbf{e}_k\mathbf{e}_k^{\mathrm{T}} \in \mathbb{R}^{N\times N}, \\
\boldsymbol{\Pi}_{\{k\}}(\boldsymbol{\alpha}) &= \begin{pmatrix}\boldsymbol{\pi}_1^k(\boldsymbol{\alpha})\mathbf{e}_k\mathbf{e}_k^{\mathrm{T}}\\ \boldsymbol{\pi}_2^k(\boldsymbol{\alpha})\mathbf{e}_k\mathbf{e}_k^{\mathrm{T}}\\ \vdots\\ \boldsymbol{\pi}_{n_c}^k(\boldsymbol{\alpha})\mathbf{e}_k\mathbf{e}_k^{\mathrm{T}}\end{pmatrix} \in \mathbb{R}^{Nn_c\times Nn_c}, \\
\mathbf{D}_{\{k\}}(\boldsymbol{\alpha}) &= \begin{pmatrix}\mathbf{D}_{\{k\}}(\boldsymbol{\pi}_1(\boldsymbol{\alpha})) & 0 & \cdots & 0\\ 0 & \mathbf{D}_{\{k\}}(\boldsymbol{\pi}_2(\boldsymbol{\alpha})) & \cdots & \vdots\\ \vdots & \vdots & & \vdots\\ 0 & \cdots & \cdots & \mathbf{D}_{\{k\}}(\boldsymbol{\pi}_{n_c}(\boldsymbol{\alpha}))\end{pmatrix},
\end{aligned}$$

and then the look-ahead model's gradient and Hessian become

$$\begin{aligned}\nabla\tilde{\mathcal{J}}_{\mathrm{CE}}^{k,\hat{y}_k}(\boldsymbol{\alpha};\mathbf{Y},\hat{\mathbf{y}}_k) =&\nabla\tilde{\mathcal{J}}_{\mathrm{CE}}(\boldsymbol{\alpha};\mathbf{Y})\\ &+\frac{1}{\gamma}\boldsymbol{\mathcal{V}}^{\mathrm{T}}\left(-\boldsymbol{\mathcal{P}}_k^{\mathrm{T}}\hat{\mathbf{y}}_k+\sum_{c=1}^{n_c}\underbrace{\frac{\mathrm{e}^{\langle\boldsymbol{\alpha},\boldsymbol{\mathcal{V}}^{\mathrm{T}}\hat{\mathbf{e}}_{k+(c-1)N}\rangle}}{\sum_{c=1}^{n_c}\mathrm{e}^{\langle\boldsymbol{\alpha},\boldsymbol{\mathcal{V}}^{\mathrm{T}}\hat{\mathbf{e}}_{k+(c-1)N}\rangle}}}_{\text{recall }=\boldsymbol{\pi}_c^k(\boldsymbol{\alpha})}\hat{\mathbf{e}}_{k+(c-1)N}\right)\\ =:&\,\nabla\tilde{\mathcal{J}}_{\mathrm{CE}}(\boldsymbol{\alpha};\mathbf{Y})+\frac{1}{\gamma}\boldsymbol{\mathcal{V}}_k^{\mathrm{T}}\left(\boldsymbol{\pi}^k-\hat{\mathbf{y}}_k\right),\end{aligned}$$

$$\begin{aligned}\nabla^2\tilde{\mathcal{J}}_{\mathrm{CE}}^{k,\hat{y}_k}(\boldsymbol{\alpha};\mathbf{Y},\hat{\mathbf{y}}_k) &= \nabla^2\tilde{\mathcal{J}}_{\mathrm{CE}}(\boldsymbol{\alpha};\mathbf{Y})+\frac{1}{\gamma^2}\boldsymbol{\mathcal{V}}^{\mathrm{T}}\Big(\mathbf{D}_{\{k\}}(\boldsymbol{\alpha})-\boldsymbol{\Pi}_{\{k\}}(\boldsymbol{\alpha})\boldsymbol{\Pi}_{\{k\}}^{\mathrm{T}}(\boldsymbol{\alpha})\Big)\boldsymbol{\mathcal{V}}\\ &=\boldsymbol{\mathcal{C}}_{\alpha}^{-1}+\frac{1}{\gamma^2}\boldsymbol{\mathcal{V}}^{\mathrm{T}}\boldsymbol{\mathcal{P}}_k^{\mathrm{T}}\Big(\mathrm{diag}\big(\boldsymbol{\pi}^k\big)-\boldsymbol{\pi}^k\big(\boldsymbol{\pi}^k\big)^{\mathrm{T}}\Big)\boldsymbol{\mathcal{P}}_k\boldsymbol{\mathcal{V}}\\ &=:\,\boldsymbol{\mathcal{C}}_{\alpha}^{-1}+\frac{1}{\gamma^2}\boldsymbol{\mathcal{V}}_k^{\mathrm{T}}\mathbf{B}_k\boldsymbol{\mathcal{V}}_k,\end{aligned}$$

where we recall $\boldsymbol{\pi}^k=\left(\boldsymbol{\pi}_1^k,\cdots,\boldsymbol{\pi}_{n_c}^k\right)^{\mathrm{T}}$ and we have defined $\boldsymbol{\mathcal{V}}_k=\boldsymbol{\mathcal{P}}_k\boldsymbol{\mathcal{V}}$ and $\mathbf{B}_k=\mathrm{diag}\big(\boldsymbol{\pi}^k\big)-\boldsymbol{\pi}^k\big(\boldsymbol{\pi}^k\big)^{\mathrm{T}}$.

With the Cholesky decomposition of the positive semi-definite $\mathbf{B}_k=\mathbf{T}_k^{\mathrm{T}}\mathbf{T}_k$, we can write

$$\begin{aligned}\tilde{\boldsymbol{\alpha}}^{k,\hat{y}_k}&=\hat{\boldsymbol{\alpha}}-\left(\nabla^2\tilde{J}_{\mathrm{CE}}^{k,\hat{y}_k}\left(\hat{\boldsymbol{\alpha}};\mathbf{Y},\hat{y}_k\right)\right)^{-1}\left(\nabla\tilde{J}_{\mathrm{CE}}^{k,\hat{y}_k}\left(\hat{\boldsymbol{\alpha}};\mathbf{Y},\hat{y}_k\right)\right)\\ &=\hat{\boldsymbol{\alpha}}-\left(\mathcal{C}_{\hat{\boldsymbol{\alpha}}}^{-1}+\frac{1}{\gamma^2}\mathcal{V}_k^{\mathrm{T}}\mathbf{T}_k^{\mathrm{T}}\mathbf{T}_k\mathcal{V}_k\right)^{-1}\left(0+\frac{1}{\gamma}\mathcal{V}_k^{\mathrm{T}}(\pi^k-\hat{\mathbf{y}}_k)\right)\\ &=\hat{\boldsymbol{\alpha}}-\frac{1}{\gamma}\mathcal{C}_{\hat{\boldsymbol{\alpha}}}\mathcal{V}_k^{\mathrm{T}}(\mathbf{I}-\mathbf{T}_k^{\mathrm{T}}(I+\mathbf{T}_k\mathbf{G}_k\mathbf{T}_k^{\mathrm{T}})^{-1}\mathbf{T}_k\mathbf{G}_k)(\pi^k-\hat{\mathbf{y}}_k),\end{aligned}$$

where we have defined $\mathbf{G}_k=\frac{1}{\gamma^2}\boldsymbol{\mathcal{V}}_k\boldsymbol{\mathcal{C}}_{\hat{\boldsymbol{\alpha}}}\boldsymbol{\mathcal{V}}_k^{\mathrm{T}}$ and applied the Woodbury Identity (A1) twice.

## Appendix C Strict Convexity of Cross-Entropy (CE) Model

We verify that the CE objective function (8) is strictly convex by showing the positive definiteness of the Hessian. The crux is merely showing that the likelihood potential for the CE model is indeed convex, per the properties of its Hessian matrix. Combining this property with the strict convexity of the graph-based regularizer proves the existence of unique minimizers of the graph-based CE objective function.

Recall objective function for the spectral truncation paradigm,

$$\tilde{\mathcal{J}}_{\mathrm{CE}}(\boldsymbol{\alpha};\mathbf{Y})=\frac{1}{2}\langle\boldsymbol{\alpha},\boldsymbol{\Lambda}_{\tau}^{\otimes}\boldsymbol{\alpha}\rangle+\sum_{j\in\mathcal{L}}\left(-\frac{1}{\gamma}\langle\mathbf{y}^j,\boldsymbol{\mathcal{P}}_j\boldsymbol{\mathcal{V}}\boldsymbol{\alpha}\rangle+\ln\left(\sum_{h=1}^{n_c}\mathrm{e}^{\langle\boldsymbol{\alpha},\boldsymbol{\mathcal{V}}^{\mathrm{T}}\hat{\mathbf{e}}_{j+(h-1)N}\rangle/\gamma}\right)\right),$$

so that by Appendix B the Hessian is

$$\nabla^2 \tilde{\mathcal{J}}_{\mathrm{CE}}(\boldsymbol{\alpha}; \mathbf{Y}) = \boldsymbol{\Lambda}_\tau^{\otimes} + \frac{1}{\gamma^2} \boldsymbol{\mathcal{V}}^{\mathrm{T}} \left( \mathbf{D}_{\mathcal{L}}(\boldsymbol{\alpha}) - \boldsymbol{\Pi}_{\mathcal{L}}(\boldsymbol{\alpha}) \boldsymbol{\Pi}_{\mathcal{L}}^{\mathrm{T}}(\boldsymbol{\alpha}) \right) \boldsymbol{\mathcal{V}}.$$

With the positive definiteness of the diagonal eigenvalue matrix $\boldsymbol{\Lambda}_\tau^{\otimes}$, we just need to show that the matrix $\nabla^2 \Phi(\boldsymbol{\alpha}; \mathbf{Y}) := \boldsymbol{\mathcal{V}}^{\mathrm{T}} \left( \mathbf{D}_{\mathcal{L}}(\boldsymbol{\alpha}) - \boldsymbol{\Pi}_{\mathcal{L}}(\boldsymbol{\alpha}) \boldsymbol{\Pi}_{\mathcal{L}}^{\mathrm{T}}(\boldsymbol{\alpha}) \right) \boldsymbol{\mathcal{V}}$ is positive semi-definite. We show that $\nabla^2 \Phi(\boldsymbol{\alpha}; \mathbf{Y}) \in \mathbb{R}^{Nn_c \times Nn_c}$ is positive semi-definite by showing that it is symmetric, with non-negative diagonal entries, and is diagonally dominant. This matrix is clearly symmetric, so we turn our attention to the other properties.

We can represent each row index $1 \leqslant r \leqslant Nn_c$ of $\nabla^2 \Phi(\boldsymbol{\alpha}; \mathbf{Y})$ via $j(r) \in \{1, 2, \cdots, N\}$ and $c(r) \in \{1, 2, \cdots, n_c\}$ such that $r = N(n_c - 1) + j$. Stated another way, for each $1 \leqslant r \leqslant Nn_c$, let $j(r) = r \bmod N$ and $c(r) = r/N + 1$. The diagonal entries then satisfy non-negativity

$$\begin{aligned} \nabla^2 \Phi(\boldsymbol{\alpha}; \mathbf{Y})_{r,r} &= \frac{1}{\gamma^2} \begin{cases} \boldsymbol{\pi}_{c(r)}^{j(r)}(1 - \boldsymbol{\pi}_{c(r)}^{j(r)}) & \text{if } j(r) \in \mathcal{L} \\ 0 & \text{else} \end{cases} \\ &= \frac{1}{\gamma^2} \delta_{j(r) \in \mathcal{L}} \boldsymbol{\pi}_{c(r)}^{j(r)}(1 - \boldsymbol{\pi}_{c(r)}^{j(r)}) \geqslant 0, \end{aligned}$$

since $\boldsymbol{\pi}_c^j \in [0, 1]$. Now, calculating diagonal dominance we have

$$\begin{aligned} \nabla^2 \Phi(\boldsymbol{\alpha}; \mathbf{Y})_{r,r} - \sum_{n \neq r}^{Nn_c} \left| \nabla^2 \Phi(\boldsymbol{\alpha}; \mathbf{y})_{r,n} \right| &= \frac{1}{\gamma^2} \delta_{j(r) \in \mathcal{L}} \left( \boldsymbol{\pi}_{c(r)}^{j(r)}(1 - \boldsymbol{\pi}_{c(r)}^{j(r)}) - \sum_{h \neq c(r)}^{n_c} \left| \mathbf{D}_{\mathcal{L}}(\boldsymbol{\pi}_c)_{j(r),j(r)} \right| \left| \mathbf{D}_{\mathcal{L}}(\boldsymbol{\pi}_h)_{j(r),j(r)} \right| \right) \\ &= \frac{1}{\gamma^2} \delta_{j(r) \in \mathcal{L}} \left( \boldsymbol{\pi}_{c(r)}^{j(r)}(1 - \boldsymbol{\pi}_{c(r)}^{j(r)}) - \boldsymbol{\pi}_{c(r)}^{j(r)} \underbrace{\sum_{h \neq c(r)}^{n_c} \boldsymbol{\pi}_h^{j(r)}}_{1 - \boldsymbol{\pi}_{c(r)}^{j(r)}} \right) \geqslant 0, \end{aligned}$$

where calculations are simple because of the diagonal structure of the matrices $\mathbf{D}_{\mathcal{L}}(\boldsymbol{\pi}_h)$ for $h \in \{1, 2, \cdots, n_c\}$. Thus, $\Phi(\boldsymbol{\alpha}; \mathbf{Y})$ is positive semi-definite, so that for all $\mathbf{x} \in \mathbb{R}^{Mn_c}$, we have

$$\langle \mathbf{x}, \boldsymbol{\mathcal{V}}^{\mathrm{T}} \nabla^2 \Phi(\boldsymbol{\alpha}; \mathbf{Y}) \boldsymbol{\mathcal{V}} \mathbf{x} \rangle = \langle \boldsymbol{\mathcal{V}} \mathbf{x}, \nabla^2 \Phi(\boldsymbol{\mathcal{V}} \boldsymbol{\alpha}; \mathbf{Y}) \boldsymbol{\mathcal{V}} \mathbf{x} \rangle \geqslant 0.$$

We therefore conclude that $\tilde{\mathcal{J}}_{\mathrm{CE}}(\boldsymbol{\alpha}; \mathbf{Y})$ is strictly convex, per the positive definiteness of its Hessian matrix, $\nabla^2 \tilde{\mathcal{J}}_{\mathrm{CE}}(\boldsymbol{\alpha}; \mathbf{Y}) = \boldsymbol{\Lambda}_\tau^{\otimes} + \nabla^2 \Phi(\boldsymbol{\alpha}; \mathbf{Y})$.

## Appendix D Adapting V-Opt and Σ-Opt

We briefly explain how we can adapt the V-Opt [26] and Σ-Opt [37] methods to fit into this spectral truncation framework (specifically the GR model) as we use it to compare against the MC acquisition function. Recall that these acquisition functions were originally derived on the HF model, with covariance matrix $\mathbf{C}_{\mathrm{HF}}$

$$\mathcal{A}_V(k) = \frac{1}{[\mathbf{C}_{\text{HF}}]_{kk}} \|[\mathbf{C}_{\text{HF}}]_{:,k}\|_2^2, \qquad \mathcal{A}_\Sigma(k) = \frac{1}{[\mathbf{C}_{\text{HF}}]_{kk}} \langle \mathbb{1}, \mathbf{C}_{\text{HF}} \mathbb{1} \rangle, \tag{D1}$$

where we note that both are functions of the look-ahead model's covariance matrix. The V-Opt criterion comes from applying the *trace* ($\text{Tr}[\cdot]$) of the look-ahead covariance $\mathbf{C}_{\text{HF}}^{+k,\hat{y}_k}$, while the $\Sigma$-Opt criterion comes from applying what is called the *survey risk* ($\langle \mathbb{1}, \cdot \mathbb{1} \rangle$) [37] to the look-ahead posterior covariance. Both of these methods are motivated by Bayesian optimal experimental design [12, 17, 48], which in the active learning context reduces to selecting unlabeled points that *minimize* these functions of the look-ahead covariance matrix. Once simplified, the acquisition functions of (D1) are then in a form to be *maximized*.

We apply similar functions to the spectral truncation modification's corresponding look-ahead posterior covariance matrix for the GR model. Recall $\boldsymbol{\alpha}|\mathbf{y} \sim \mathcal{N}(\hat{\boldsymbol{\alpha}}, \mathbf{C}_{\hat{\boldsymbol{\alpha}}})$ where $\mathbf{C}_{\hat{\boldsymbol{\alpha}}} = \left(\boldsymbol{\Lambda}_\tau + \frac{1}{\gamma^2}\mathbf{V}^{\mathrm{T}}\mathbf{P}^{\mathrm{T}}\mathbf{P}\mathbf{V}\right)^{-1}$ and $\hat{\boldsymbol{\alpha}} = \frac{1}{\gamma^2}\mathbf{C}_{\hat{\boldsymbol{\alpha}}}\mathbf{V}^{\mathrm{T}}\mathbf{P}^{\mathrm{T}}\mathbf{y}$ so that we have $\mathbf{u} \sim \mathcal{N}(\mathbf{V}\hat{\boldsymbol{\alpha}}, \mathbf{V}\mathbf{C}_{\hat{\boldsymbol{\alpha}}}\mathbf{V}^{\mathrm{T}})$. We compute the V-Opt and $\Sigma$-Opt acquisition functions as

$$\begin{aligned}
\text{Tr}\left[\mathbf{V}\mathbf{C}_{\hat{\boldsymbol{\alpha}}}^{+k,\hat{y}_k}\mathbf{V}^{\mathrm{T}}\right] &= \text{Tr}\left[\mathbf{V}\left(\mathbf{C}_{\hat{\boldsymbol{\alpha}}} - \frac{1}{\gamma^2 + \mathbf{v}_k^{\mathrm{T}}\mathbf{C}_{\hat{\boldsymbol{\alpha}}}\mathbf{v}_k}\mathbf{C}_{\hat{\boldsymbol{\alpha}}}\mathbf{v}_k\mathbf{v}_k^{\mathrm{T}}\mathbf{C}_{\hat{\boldsymbol{\alpha}}}\right)\mathbf{V}^{\mathrm{T}}\right] \\
&= \text{const} - \frac{1}{\gamma^2 + \mathbf{v}_k^{\mathrm{T}}\mathbf{C}_{\hat{\boldsymbol{\alpha}}}\mathbf{v}_k}\|\mathbf{C}_{\hat{\boldsymbol{\alpha}}}\mathbf{v}_k\|_2^2, \\
\langle \mathbb{1}, \mathbf{V}\mathbf{C}_{\hat{\boldsymbol{\alpha}}}^{+k,\hat{y}_k}\mathbf{V}^{\mathrm{T}}\mathbb{1}\rangle &= \langle \mathbb{1}, \mathbf{V}\left(\mathbf{C}_{\hat{\boldsymbol{\alpha}}} - \frac{1}{\gamma^2 + \mathbf{v}_k^{\mathrm{T}}\mathbf{C}_{\hat{\boldsymbol{\alpha}}}\mathbf{v}_k}\mathbf{C}_{\hat{\boldsymbol{\alpha}}}\mathbf{v}_k\mathbf{v}_k^{\mathrm{T}}\mathbf{C}_{\hat{\boldsymbol{\alpha}}}\right)\mathbf{V}^{\mathrm{T}}\mathbb{1}\rangle \\
&= \text{const} - \frac{1}{\gamma^2 + \mathbf{v}_k^{\mathrm{T}}\mathbf{C}_{\hat{\boldsymbol{\alpha}}}\mathbf{v}_k}\langle \mathbf{V}^{\mathrm{T}}\mathbb{1}, \mathbf{C}_{\hat{\boldsymbol{\alpha}}}\mathbf{v}_k\rangle^2,
\end{aligned}$$

where we have used the orthonormality of the columns of $V$ and the Woodbury Matrix Identity (A1). Now as both the V-Opt and $\Sigma$-Opt acquisition functions were originally formulated to *minimize* these functions (respectively, $\text{Tr}[\cdot], \langle \mathbb{1}, \cdot \mathbb{1}\rangle$) of the posterior covariance matrix, we can rewrite these modified acquisition functions in the *maximizing* paradigm

$$\mathcal{A}_V(k) := \frac{1}{\gamma^2 + \mathbf{v}_k^{\mathrm{T}}\mathbf{C}_{\hat{\boldsymbol{\alpha}}}\mathbf{v}_k}\|\mathbf{C}_{\hat{\boldsymbol{\alpha}}}\mathbf{v}_k\|_2^2, \qquad \mathcal{A}_\Sigma(k) := \frac{1}{\gamma^2 + \mathbf{v}_k^{\mathrm{T}}\mathbf{C}_{\hat{\boldsymbol{\alpha}}}\mathbf{v}_k}\langle \mathbb{1}, \mathbf{C}_{\hat{\boldsymbol{\alpha}}}\mathbf{v}_k\rangle^2. \tag{D2}$$

**Acknowledgements** Kevin Miller was supported by the DOD National Defense Science and Engineering Graduate (NDSEG) Research Fellowship. Andrea Bertozzi is supported by the NGA under Contract No. HM04762110003. Any opinions, findings and conclusions or recommendations expressed in this material are those of the authors and do not necessarily reflect the views of the NGA. Approved for public release, 22-057.

**Author Contributions** ALB oversaw the research including the general area of active learning and the application problems of hyperspectral imagery. She also contributed to the writing of the manuscript. KSM conceived the Model Change acquisition function and its mathematical foundation. He also wrote the Python code, conducted the numerical experiments, and contributed to the writing of the manuscript.

**Funding** This work was supported by the DOD National Defense Science and Engineering Graduate (NDSEG) Research Fellowship and the NGA under Contract No. HM04762110003. Any opinions, findings and conclusions or recommendations expressed in this material are those of the authors and do not necessarily reflect the views of the NGA. Approved for public release, 22-057.

**Availability of Data and Materials** Not applicable.

**Code Availability** The code to reproduce experiments is available at https://github.com/millerk22/model-change-paper.

## Compliance with Ethical Standards

**Conflict of Interest** On behalf of all authors, the corresponding author states that there is no conflict of interest.

**Ethics Approval** Not applicable.

**Consent to Participate** Not applicable.

**Consent for Publication** Not applicable.